%% file: main.tex
\documentclass[10pt,twocolumn,letterpaper]{article}

\usepackage[pagenumbers]{cvpr} 

\usepackage{wrapfig}
\usepackage{graphicx}
\usepackage{booktabs}
\usepackage{multirow}
\usepackage{tabularray}
\usepackage[innercaption]{sidecap}
\usepackage{wrapfig}
\usepackage{colortbl}
\usepackage[accsupp]{axessibility}  
\usepackage{float}
\usepackage{makecell}
\usepackage{caption}
\usepackage{stfloats}
\usepackage{soul}
\usepackage{xcolor}
\sethlcolor{cyan!15}
\definecolor{cvprblue}{rgb}{0.21,0.49,0.74}
\usepackage[pagebackref,breaklinks,colorlinks,allcolors=cvprblue]{hyperref}

\def\paperID{} 
\def\confName{CVPR}
\def\confYear{2026}

\title{\vspace{-2mm}SketchDeco: Training-Free Latent Composition for Precise Sketch Colourisation\vspace{-4mm}}

\author{
\href{https://chaitron.github.io/}{Chaitat Utintu}   \hspace{.4cm}
\href{https://www.surrey.ac.uk/people/yi-zhe-song}{Yi-Zhe Song}\\
SketchX, CVSSP, University of Surrey.\\
{\texttt{\{chaitat.u, y.song\}@surrey.ac.uk}}\\
\vspace{-1mm}
}

\begin{document}
\twocolumn[{%
\renewcommand\twocolumn[1][]{#1}%
\maketitle
    \captionsetup{type=figure}
    \vspace{-0.6cm}
    \includegraphics[width=\linewidth]{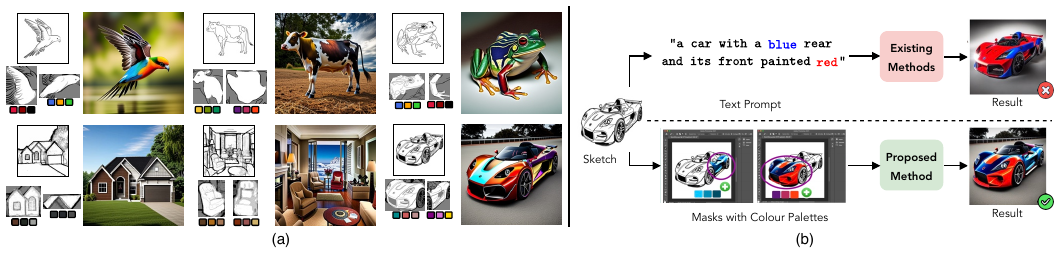} \\[-0.7cm]   
    \caption{(a) Set of results generated by the proposed method. (b) Prior works on user-guided sketch colouring \cite{diff4} that use text descriptions often lack precision. Hence, we focus on precision and convenience by requiring designers to specify the region mask (\eg, use Photoshop) and its desired colour palettes. Our training-free method finds the optimal way to colour specified regions with chosen colours.
    }
    \label{fig:teaser}
    \vspace{3mm}
}]

\begin{abstract}
We introduce \texttt{SketchDeco}, a training-free approach to sketch colourisation that bridges the gap between professional design needs and intuitive, region-based control. Our method empowers artists to use simple masks and colour palettes for precise spatial and chromatic specification, avoiding both the tediousness of manual assignment and the ambiguity of text-based prompts. We reformulate this task as a novel, training-free \textit{composition} problem. Our core technical contribution is a guided latent-space blending process: we first leverage diffusion inversion to precisely ``paint'' user-defined colours into specified regions, and then use a custom self-attention mechanism to harmoniously blend these local edits with a globally consistent base image. This ensures both local colour fidelity and global harmony without requiring any model fine-tuning. Our system produces high-quality results in 15--20 inference steps on consumer GPUs, making professional-quality, controllable colourisation accessible.
\end{abstract}

\vspace{-4mm}
\section{Introduction}
\label{sec:intro}

Transforming a line-art sketch into a fully coloured image is a fundamental task in creative workflows, from animation storyboards and product design to concept art. This process requires not just the artistic application of colour, but also precise spatial control to define regions and maintain coherence. While large-scale diffusion models \cite{ldm, saharia2022imagegen} have revolutionized image synthesis, achieving this fine-grained, region-specific \textit{colour} control remains a major challenge. Text prompts are semantically powerful but spatially ambiguous. And while structural controls like sketches can define form \cite{diff4}, they do not solve the core problem of how to grant artists direct command over \textit{where} and \textit{how} colour is applied without compromising the model's rich generative priors.

The challenge is a trade-off between control and quality. Traditional methods, from manual assignment \cite{zhang2018twostage, sangkloy2017scribbler} to reference-based transfer \cite{yan2024colorizediffusion, ref1}, offer precision but are far too labour-intensive for modern workflows. Conversely, text-guided approaches \cite{text1, text3} are fast but imprecise, often resulting in semantic errors and ``colour bleeding'' that require manual correction.

This paper argues that the solution is not more training, but a novel, training-free \textit{composition} framework. We introduce \texttt{SketchDeco}, a system that reformulates region-controlled colourisation as a latent-space blending problem. We empower artists with simple, practical inputs -- a sketch, region masks, and colour palettes -- and integrate them \textit{without} fine-tuning.

Our core intuition is to separate global consistency from local control. Instead of trying to force a single model to do both at once, our system first generates a globally consistent, fully-coloured base image from the sketch. Then, it \textit{composites} the user's specific colour choices (from palettes) into their specified regions (from masks) at the latent level. This approach allows us to blend the local edits harmoniously with the global base image, rather than fighting against the model's priors.

More specifically, to achieve this composition, we employ a \textit{guided inversion and blending} process. We leverage diffusion inversion \cite{denoise, mokady2022nulltext} with DPM-Solver++ \cite{lu2023dpmsolver} to find an optimal noise vector and precisely ``paint'' the user's colours into the target latent regions. To ensure this edit is both precise and stable, we use guided sampling with classifier-free guidance (CFG) \cite{ho2021classifier-free} and an exceptional prompt \cite{tficon}. Finally, a custom self-attention mechanism, modulated by a scaling factor $\tau$, harmoniously blends these local edits, preserving the original sketch structure while ensuring global coherence.

Our unified, training-free approach provides the following contributions:
\begin{itemize}
    \item A novel region-guided framework that composes masks and palettes for precise, flexible colour control.
    \item A latent-space composition technique combining diffusion inversion and guided sampling to ensure both local colour fidelity and global harmony.
    \item A custom attention mechanism that adaptively balances sketch structure preservation with controlled colour diffusion.
    \item An efficient, training-free method that produces high-quality results in 15-20 steps on consumer GPUs.
\end{itemize}

\vspace{-1mm}
\section{Related Works}
\label{sec:related}
\vspace{-1.5mm}

    \noindent \textbf{Sketch-to-Image Generation:} Sketch-to-image generation \cite{koley2023picture} diverges from traditional image-to-image translation \cite{i2i1,i2i2,i2i3,i2i4,i2i5,i2i6} to generate realistic images from input sketches, bridging the gap between hand-drawn sketches and real images. SketchyGAN \cite{sketchygan} pioneers this task by employing GANs \cite{gan} alongside edge-preserving image augmentations, while ContextualGAN \cite{contextualgan} utilises conditional GANs to learn joint image-sketch representations. Given the difficulty in acquiring paired sketch-image datasets, some researchers have explored unsupervised \cite{unsupervise} and self-supervised \cite{selfsupervise, sain20223T} learning paradigms to tackle this challenge. Subsequent studies have enhanced results by fine-tuning pretrained StyleGAN models \cite{gan1} or leveraging VQGAN to minimise discrepancies between ground truth image-sketch embeddings \cite{gan2}. Despite these advancements, existing GANs-based methods for sketch-to-image generation often suffer from issues of unstable training and inferior quality, leading to a shift towards diffusion models \cite{diffusion,diff2,diff3,diff4,gligen,t2iadapter}. In this paper, we utilise pretrained text-to-image (T2I) diffusion models and address sketch-to-image generation problem 
    as a downstream task.
    \vspace{2mm}

    \noindent \textbf{Diffusion Models for Vision-based Tasks:}  
    Diffusion models \cite{diffusion,denoise,thermo}, a recent class of generative models, outperform GANs \cite{gan} and VAEs \cite{vae} in both output quality and training stability by learning to reverse a Markov process through a denoising autoencoder. To improve efficiency, Rombach \etal \cite{ldm} introduced Latent Diffusion Models (LDMs), later popularised as Stable Diffusion (SD), which perform the diffusion process in a latent space rather than in pixel space, significantly reducing computational costs while enabling conditional control via cross-attention \cite{transformer}. Text-to-image (T2I) diffusion models employ textual inputs to iteratively guide denoising, achieving state-of-the-art results in applications such as image synthesis \cite{diff1}, style transfer \cite{user3}, content manipulation \cite{cyclenet}, inpainting \cite{repaint}, and compositional generation \cite{tficon}. Beyond text-based conditioning, additional modalities, such as sketches, enhance control and fidelity; however, naive methods necessitate full model retraining \cite{diff3,diff1,user3}, leading to substantial computational overhead. To mitigate this, techniques like ControlNet \cite{diff4}, GLIGEN \cite{gligen}, and T2I-Adapter \cite{t2iadapter} fine-tune hypernetworks for domain adaptation. More recent approaches introduce conditioning mechanisms that circumvent the need for retraining or fine-tuning \cite{freeu,tficon,freecontrol}, refining the UNet's \cite{unet} denoising process by leveraging cross-attention maps \cite{transformer}. In this work, we adopt a training-free methodology to address the realistic sketch colourisation problem, ensuring efficiency while preserving fidelity and control in the generated outputs.


    \vspace{2mm}

    \noindent \textbf{Image and Sketch Colourisation:} Deep learning (DL)-based image colourisation has advanced well beyond traditional methods \cite{trad_color1,trad_color2}, with approaches commonly categorised as reference-based \cite{ref1,deep5,ref4,ref5,ref6}, text-based \cite{text1,text2,text3,text4,text5}, or user-guided \cite{text2,user1,user2,user3,user4}.
Reference-based methods transfer colour via pixel or semantic correspondences \cite{ref4,ref5}, often using GANs \cite{deep5} or pretrained networks \cite{ref6}, but remain prone to semantic misalignment.
Text-based methods employ language cues through attention \cite{text1}, palettes \cite{text2}, or large-scale models \cite{text3}, yet suffer from ambiguity and limited control.
User-guided methods spread user-specified hints \cite{text2,bhunia2023saliency}, optimise colour propagation \cite{user1}, or combine global \cite{user2,user3} and local inputs \cite{user4,bhunia2022doodle}, though spatial coherence and intuitive control remain challenging. Sketch colourisation \cite{zhuang2025cobra,liu2025manganinja, zhuang2025colorflow, yan2024colorizediffusion,wu2023flexicon,wu2023self} extends this field, sharing similar principles but facing distinct difficulties due to sparse structure and lack of texture. As such, it remains a relatively new and challenging area with considerable potential for further exploration. In this study, we introduce a user-guided, text-free framework that enhances creative expression by leveraging a sketch, region-controlled masks, and corresponding colour palettes, thereby enabling precise, coherent, and spatially controllable sketch colourisation.

\vspace{-1.5mm}
\section{Proposed Method}
\label{sec:proposed}
\vspace{-1.5mm}

\begin{figure}[h!]
    \centering
        \includegraphics[width=\linewidth]{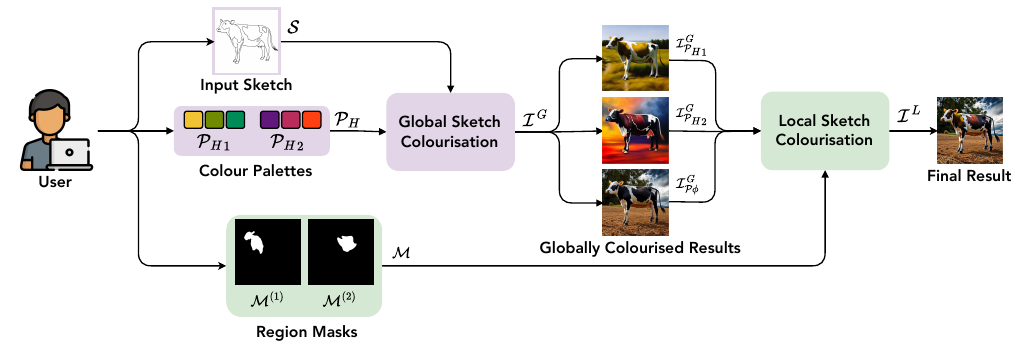}
        \vspace{-7mm}
    \caption{\textbf{Framework Overview.} Given an input sketch and region-specific colour palettes with corresponding masks, our method employs a divide-and-conquer strategy consisting of two sequential stages: \textbf{Global} and \textbf{Local Sketch Colourisations}.}
    \label{fig:overview}
    \vspace{-4mm}
\end{figure}

\noindent \textbf{Overview:} We introduce a novel training-free paradigm for sketch colourisation, as illustrated in Fig.~\ref{fig:overview}. Our method runs efficiently on a consumer-grade GPU and leverages the rich prior knowledge embedded in large-scale diffusion models \cite{ldm}. Given an input sketch \(\mathcal{S} \in \mathbb{R}^{512\times 512\times3}\), a set of colour palettes \( \left\{ {\mathcal{P}_{H}}_{1}, {\mathcal{P}_{H}}_{2}, ..., {\mathcal{P}_{H}}_{n}\right\} \), where \(\mathcal{P_{H}}\) comprises hexadecimal colour codes, and corresponding region masks \(\left\{\mathcal{M}^{(1)}, \mathcal{M}^{(2)}, \dots, \mathcal{M}^{(n)}\right\}\), with each \(\mathcal{M} \in \mathbb{R}^{512\times 512\times1}\), we adopt a divide-and-conquer strategy consisting of two sequential stages: \textit{global sketch colourisation} (Sec.~\ref{subsec:global}) and \textit{local sketch colourisation} (Sec.~\ref{subsec:local}). The global stage generates multiple plausible colourised outputs \(\mathcal{I}^{G}\) that preserve both the structure of \(\mathcal{S}\) and the specified colour palettes \(\mathcal{P_{H}}\). The local stage refines these outputs using region masks \(\mathcal{M}\), ensuring seamless composition and smooth colour transitions. The final result, \(\mathcal{I}^{L}\), preserves spatial coherence while satisfying user-defined constraints.

\vspace{-3mm}

    \subsection{Background}
    \vspace{-1mm}
    \noindent \textbf{Pixel Space Diffusion Models:} Diffusion Probabilistic Models (DPMs) \cite{diffusion,dhariwal2021diffusionbeatsGAN,denoise} approximate a data distribution \( p(x) \) by iteratively denoising a Gaussian-distributed variable. The denoising function is trained to reverse a fixed Markov Chain process of length \( T \). Diffusion modeling consists of two stochastic processes: \( forward \) and \( backward \) diffusion \cite{denoise}. In training, \( forward \) diffusion gradually corrupts an image \( x_{0}\in \mathbb{R}^{H\times W\times3} \) by adding Gaussian noise, producing \( x_{t} \) as \( x_{t}=\sqrt{ \bar{\alpha}_{t}}x_{0} + (\sqrt{1-\bar{\alpha}_{t}})\epsilon \), where, \( \alpha_{t} \) controls noise levels from \( \alpha_{0}=1 \) to \( \alpha_{T}\simeq 0 \), and \( t \) is sampled from \( \{1,...,T\} \) \cite{denoise}. The \( backward \) process employs denoising autoencoders \( \epsilon_{\theta }(\cdot) \) to predict clean images by minimizing:
        \vspace{-2mm}
        {\small
        \begin{equation}
            \mathcal{L}_\textrm{DM}=\mathbb{E}_{x,\epsilon \sim \mathcal{N}(0,1),t}\left [ \| \epsilon -\epsilon _{\theta }(x_{t},t) \| _{2}^{2}\right ].
            \vspace{-2mm}
        \end{equation}}
    During inference, the trained network \( \epsilon_{\theta }(\cdot) \) iteratively refines a Gaussian noise sample \( x_{T} \) over \( T \) steps to reconstruct \( x_{0} \), recovering the data distribution \cite{denoise}.

    \noindent \textbf{Latent Diffusion Models:} Latent Diffusion Models (LDMs, e.g., Stable Diffusion \cite{ldm}) learn data distributions in a compact latent space rather than directly in pixel space. An autoencoder with encoder \(\mathcal{E}(\cdot)\) and decoder \(\mathcal{D}(\cdot)\) (UNet backbone \cite{unet}) performs diffusion efficiently \cite{ldm}. Given an image \(x_{0}\in \mathbb{R}^{H\times W\times3}\), the encoder downsamples it by \(f=H/h=W/w\) into \(z_{0}\in \mathbb{R}^{h\times w\times d}\) as \(z_{0}=\mathcal{E}(x_{0})\), and the decoder reconstructs \(\tilde{x_{0}}=\mathcal{D}(z_{0})=\mathcal{D}(\mathcal{E}(x_{0}))\). Conditional generation is enabled by adding cross-attention \cite{transformer} so that the denoising network \(\epsilon_{\theta}(z_{t},t,c)\) learns \(p(z|c)\), where \(c\) is the condition. The training objective is:
\vspace{-2mm}
{\small
\begin{equation}
\label{eq2}
    \mathcal{L}_\textrm{LDM}=\mathbb{E}_{\mathcal{E}(x),c,\epsilon \sim \mathcal{N}(0,1),t}\left [ \left \| \epsilon -\epsilon _{\theta }(z_{t},t,c) \right \| _{2}^{2}\right ]
    \vspace{-1mm}
\end{equation}   }

LDMs support various conditions (\eg, text \cite{ldm}, sketch \cite{diff4}) using the same objective. For text conditioning, a prompt \(p\) is embedded via \(\textbf{T}(\cdot)\) and encoded by the pretrained CLIP text encoder \(\mathrm{\mathcal{E}}_{t}(\cdot)\) \cite{clip} to obtain \(t_{p}=\mathrm{\mathcal{E}}_{t}(\textbf{T}(p))\). The text encoding is injected into UNet layers through cross-attention \cite{transformer}, defined as \(\text{Attention}(Q,K,V)=\text{softmax}\left(\frac{QK^{T}}{\sqrt{d}}\right)\cdot V\), with
    \vspace{-2mm}
    {\small
    \begin{equation}
        Q=W_{Q}\cdot z_{t},\, 
        K=W_{K}\cdot t_{p},\, 
        V=W_{V}\cdot t_{p}
        \vspace{-2mm}
    \end{equation}}
where \(W_{Q}, W_{K}, W_{V}\) are learnable projection matrices; \(W_{K}, W_{V}\) linearly project \(t_{p}\in \mathbb{R}^{77\times768}\) to Keys and Values, and \(W_{Q}\) projects noisy latents to Query maps \cite{ldm}.

\begin{figure*}[t]
\centering
\vspace{-7mm}
\includegraphics[width=\textwidth]{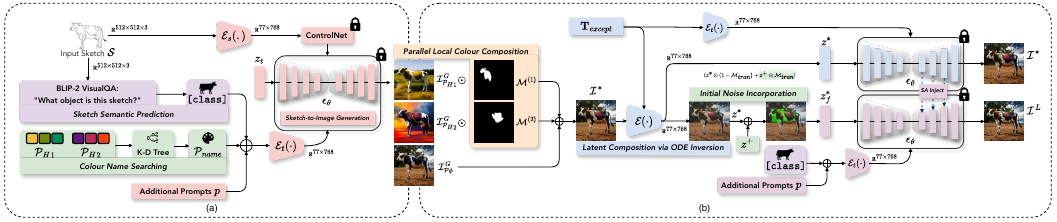}
\vspace{-7mm}
\caption{\textbf{Global and Local Sketch Colourisation Stages.} (a) In \textbf{global stage}~\ref{subsec:global}, given a sketch \(\mathcal{S}\) and colour palettes \(\{\mathcal{P_H}\}_{i=1}^{n}\), BLIP-2 \cite{blip2} infers sketch class semantics, a K-D Tree \cite{kdtree} maps palette hexcodes to colour names, and Scribble ControlNet \cite{diff4} generates globally colourised results \(\{\mathcal{I}^{G}_{\mathcal{P_H}_{i}}\}_{i=1}^{n}\) and an auxiliary image \(\mathcal{I}^{G}_{\mathcal{P}\phi}\), preserving \(id(\mathcal{S})\) and \(colour(\mathcal{P_{H}})\), while enabling interactive refinement for adjusting unsatisfying textures further.
 (b) In \textbf{local stage}~\ref{subsec:local}, pre-colourised regions from \(\{\mathcal{I}^{G}_{\mathcal{P_H}_{i}}\}_{i=1}^{n}\) are composed with background from \(\mathcal{I}^{G}_{\mathcal{P}\phi}\) to form composited image \(\mathcal{I}^{\ast}\), which is then inverted to noisy latents \(z^{\ast}\) and refined via guided-sampling \cite{lu2023dpmsolver}. Noise incorporation enhances boundary smoothness, while SA injection, guided by \(\tau\), produces the final result \(\mathcal{I}^{L}\) with smooth blending and structural fidelity.}
\label{fig:method}
\vspace{-5mm}
\end{figure*}

    \subsection{Global Sketch Colourisation}
    \label{subsec:global}

    \vspace{-1mm}
    We present a novel user-guided global sketch colourisation framework (see Fig.~\ref{fig:method} a) operating on input sketch \(\mathcal{S}\in \mathbb{R}^{512\times512\times3}\) and colour palette \(\mathcal{P_{H}}=\left\{ h\;|\;h\;\text{is a colour hexcode}\right\}\). The model leverages generative priors from pretrained Stable Diffusion to achieve two goals: \textit{(a)} generate faithful images \(\mathcal{I}^{G}\) preserving sketch identity and structure (\ie \(id(\mathcal{I}^{G}) \approx id(\mathcal{S})\)), and \textit{(b)} produce vivid colours aligned with \(\mathcal{P_{H}}\) (\ie \(\textit{colour}(\mathcal{I}^{G}) \approx \textit{colour}(\mathcal{P_{H}})\)). The main challenges are: \textit{(i)} the prompt-free setup hinders sketch-to-image generation without the sketch’s class label in textual conditioning \cite{diff4}, and \textit{(ii)} colour hexcodes must be mapped to standard colour names.

    To address these, we use BLIP-2 \cite{blip2} in a VQA setup to infer sketch semantics, a K-Dimensional Tree (K-D Tree) \cite{kdtree} to match hexcodes with colour names, and a pretrained Scribble ControlNet \cite{diff4} for structure-preserving sketch-to-image generation. Building on \cite{cong2024imagination}, our pipeline directly processes the input sketch, integrates BLIP-2 class cues, and employs K-D Tree colour retrieval for enabling training-free, user-guided sketch colourisation.

    \noindent \textbf{Sketch Semantic Prediction:} We utilise BLIP-2 \cite{blip2}, a vision–language pre-training framework that effectively aligns image and text modalities via its Q-former module \cite{blip2}. The Q-former connects CLIP’s image encoder with a large language model (LLM) through learnable query embeddings that attend to relevant visual regions. We adopt BLIP-2’s Visual Question Answering (VQA) setup to infer class semantics from input sketches by prompting (\eg, ``\textit{What is the object in this sketch?}''). This approach provides strong zero-shot generalisation across diverse sketch types, outperforming previous works \cite{zhang2018twostage, yan2024colorizediffusion, kim2023diffblender} that relied solely on task-specific sketch encoders.

\begin{figure}[hbt!]
    \centering
        \includegraphics[width=.95\linewidth]{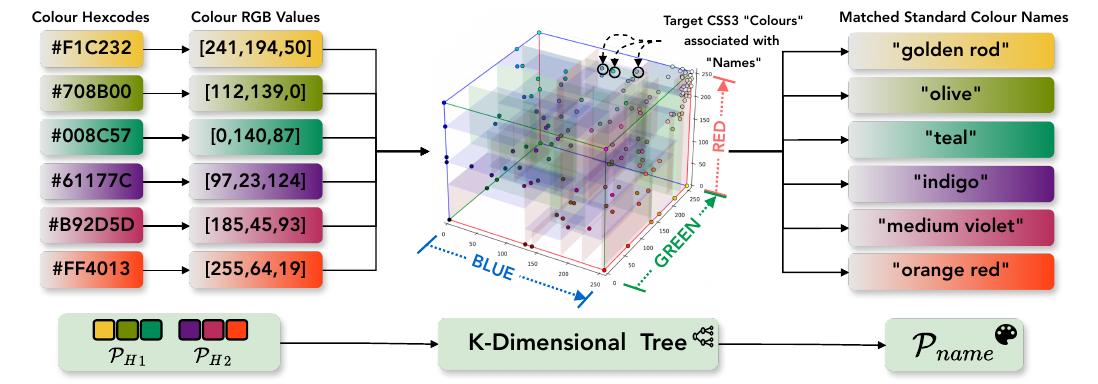}
        \vspace{-1mm}
    \caption{3D search space constructed using K-D Tree algorithm.
}
    \label{fig:kdtree}
    \vspace{-2mm}
\end{figure}

    \noindent \textbf{Colour Name Searching:} We employ the CSS3 colour database \cite{css3-colours} and a K-Dimensional Tree (K-D Tree) \cite{kdtree} to efficiently map colour hexcodes to their nearest named counterparts in RGB space. A K-D Tree with \(K=3\) is constructed from the \(147\) CSS3 reference colours (see Fig.~\ref{fig:kdtree}), ensuring that any colour in \(\mathbb{R}^{3}\) can be associated with a valid name. Every colour hexcodes \(\mathcal{P_H}=\{h_i\}_{i=1}^{n}\) are converted to RGB values, and the tree retrieves the closest colour names \(\mathcal{P}_{name}=\{c_s^i\}_{i=1}^{n}\). The retrieved names are integrated into textual prompts that incorporate class semantics and descriptive cues. The final guidance \(t_p\) is formatted as: ``\texttt{[class]}\textit{, hyper-realistic, quality, photography style, using only colours in colour palette of} \texttt{[$c_s^1$]}, - \texttt{[$c_s^2$]} - $\dots$ \texttt{[$c_s^n$]}'' and negative prompts ``\textit{drawing look, sketch look, ...}''. See supplement for more details about the K-D Tree algorithm (e.g., design choices, and advantages over LLMs).

\vspace{-3mm}

    \vspace{-1mm}
    \noindent \textbf{Sketch-to-Image Generation:}
    To generate a realistic image \(\mathcal{I}^{G}\) based on input sketch \(\mathcal{S}\), we employ the generative prior derived from a robust pretrained T2I diffusion model. Unlike textual conditioning \(t_{p}\), sketches embody spatial information \cite{sangkloy2022textsketch}. Thus, we harness the capabilities of the pretrained Scribble ControlNet \cite{diff4} to attain fidelity and realism in the sketch-to-image generation. Given the sketch latents \(z_{s} = \mathcal{E}_{s}(\mathcal{S}); \ z_{s} \in \mathbb{R}^{h \times w \times d}\), the noise estimator \(\epsilon _{\theta }(z_{t},t,t_{p},z_{s})\) needs to incorporate sketch condition \(p(z|t_{p},z_{s})\). In this step, we generate \(n+1\) images: the globally colourised results \(\{\mathcal{I}^{G}_{\mathcal{P_H}_{i}}\}_{i=1}^{n}\), derived from the given colour palettes \(\{{\mathcal{P_H}_{i}}\}_{i=1}^{n}\), and an auxiliary image \(\mathcal{I}^{G}_{\mathcal{P}\phi}\), which has no colour description (see Fig.~\ref{fig:method}). This extra image serves as the background for composition steps.

\vspace{1mm}

\begin{figure}[hbt!]
        \centering
            \includegraphics[width=.9\linewidth]{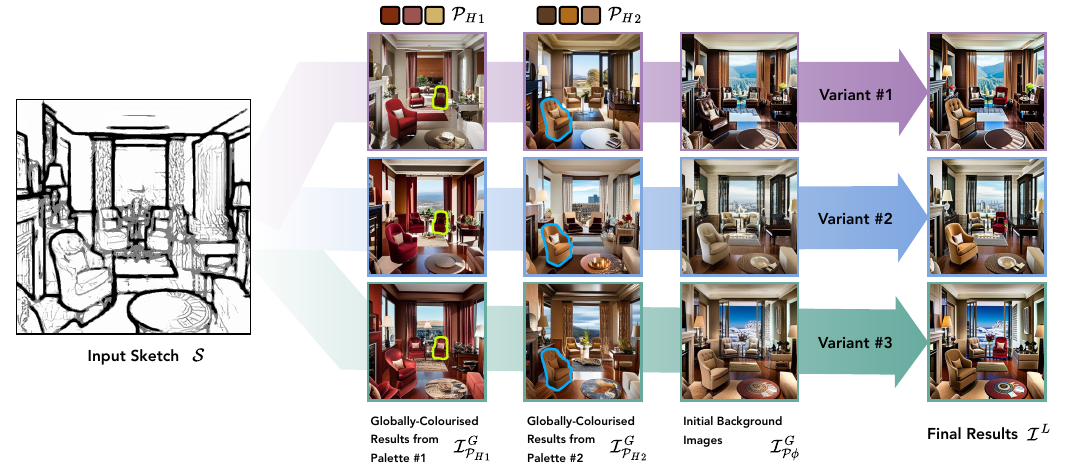}
            \vspace{-3mm}
        \caption{User-interactive refinement process for interior design.}
        \label{fig:user_interact}
        \vspace{-5mm}
    \end{figure}

    \noindent \textbf{User-interactive Refinement:} Users can apply local colour composition in the latent space \(z\) using user-defined masks \(\{\mathcal{M}^{(i)}\}_{i=1}^{n}\). However, since interactions in latent space are not directly perceptible to the human eye, this approach offers limited controllability and interpretability for creative workflows, where human feedback is essential \cite{cong2024imagination}. To address this, our pipeline renders global previews (\ie, \(\{\mathcal{I}^{G}_{\mathcal{P_H}_{i}}\}_{i=1}^{n}\) and \(\mathcal{I}^{G}_{\mathcal{P}\phi}\)) in pixel space, allowing users to inspect, regenerate, and compare variants with different random seeds (see Supp. video). Rather than explicitly compositing latent features, our method initiates composition in pixel space, then applies ODE inversion to blend the results back into latent space during the local stage (\cref{fig:method} b), implicitly reproducing the same compositional effect. This hybrid approach retains the flexibility of latent editing while enabling direct, interpretable interaction ideal for creative domains, like interior design, where rapid exploration of colour-texture combinations is critical (see Fig.\ref{fig:user_interact}).

\vspace{-2mm}

    \subsection{Local Sketch Colourisation}
    \label{subsec:local}
    \vspace{-1mm}

    In local sketch colourisation (see Fig.~\ref{fig:method} b), we are, to the best of our knowledge, the first to reframe this task as a combination of \textit{image composition} \cite{tficon} and \textit{reconstruction} \cite{lu2023dpmsolver}. For text-guided approach \cite{diff4}, textual prompts often struggle to provide precise colour control (see Fig.~\ref{fig:teaser} b), so we instead introduce spatial colour information directly into the noisy latent space rather than relying solely on the cross-attention mechanism \cite{transformer}. In our approach, pre-colourised regions derived from \({\mathcal{I}}^{G}_{\mathcal{P}_{H}}\) are composed with \({\mathcal{I}}^{G}_{\mathcal{P}\phi}\). Since all global results \({{\mathcal{I}}^{G}}\) originate from the same sketch, this composition step inherently preserves sketch faithfulness. The composited image is then inverted back into the latent space, \textit{allowing the colour information to be directly embedded in the noisy latents}, and finally performs latent composition to refine boundaries and ensure coherent visual blending, all without any model retraining or fine-tuning.

    Therefore, the final colourised image \(\mathcal{I}^{L}\) aims to satisfy three objectives: (a) preserve the identity of the input sketch (\ie, \(id(\mathcal{I}^{L}) \approx id(\mathcal{S})\)), (b) maintain local colour consistency within each mask region \(\mathcal{M}\) such that the resulting colours align with the provided palette \(\mathcal{P_{H}}\) (\ie, \(colour(\mathcal{I}^{L}\odot\mathcal{M})\approx colour(\mathcal{P_{H}})\)), and (c) ensure smooth and coherent transitions between locally coloured regions.

\vspace{2mm}
    \noindent \textbf{Parallel Local Colour Composition:}
Since \(\mathcal{I}^{G}\) inherits both \(id(\mathcal{S})\) and \(colour(\mathcal{P_{H}})\), we first apply each mask \(\mathcal{M}^{(i)}\) to the corresponding \({\mathcal{I}}^{G}_{{\mathcal{P_{H}}}_{i}}\) to isolate its region of interest: \({\mathcal{I}}^{G}_{\mathcal{M}^{(i)}} = {\mathcal{I}}^{G}_{{\mathcal{P_{H}}}_{i}} \odot \mathcal{M}^{(i)}, i = 1, \ldots, n,\) where \(\odot\) denotes the Hadamard product. The composited image \({\mathcal{I}}^{\ast}\) is then obtained as: \({\mathcal{I}}^{\ast} = {\mathcal{I}}^{G}_{\mathcal{P}\phi} \odot (1 - {\mathcal{M}}_{tot}) + \sum_{i=1}^{n} {\mathcal{I}}^{G}_{\mathcal{M}^{(i)}},\) with \\\({\mathcal{M}}_{tot}=\max_{i=1}^{n}{\mathcal{M}}^{(i)}\).
This design enables \textit{parallel processing of multiple masks}, where the number of regions only affects the number of intermediate globally colourised results \({\mathcal{I}}^{G}_{\mathcal{P}_{H}}\), without impacting the quality of \({\mathcal{I}}^{\ast}\) and \(\mathcal{I}^{L}\).

\vspace{2mm}
    \noindent \textbf{Latent Composition via ODE Inversion:} The composited image $\mathcal{I}^{\ast}$ satisfies our second objective -- local colour consistency -- but lacks smooth transition between local coloured regions (third objective). To achieve smooth transitions between local colourised regions, a crucial step involves transforming the composited image \({\mathcal{I}}^{\ast}\) into a latent representation \(z^{\ast}\). The sampling process within diffusion probabilistic models (DPM) can be executed by solving ordinary differential equations (ODE) from \(T\) to \(0\) \cite{song2021scorebased}: 
    \vspace{-2mm}
    {\small
    \begin{equation}
    \label{eq:ode}
    \hspace{-2mm}\frac{dx_{t}}{dt}=f(t)x_{t}+\frac{g^{2}(t)}{2\sigma_{t}}\epsilon_{\theta}(x_{t},t),\;\text{ where, }\;\;x_{T}\sim \mathcal{N}(0,\tilde{\sigma}^{2}I),
    \vspace{-2mm}
    \end{equation}}
    
    \noindent To perform the inversion process, we integrate the diffusion ODE from (0) to (T) using DPM-Solver++\cite{lu2023dpmsolver}, chosen for its efficiency in generating high-quality samples within 15–20 steps, compared with the 100–250 steps typically required by DDIM\cite{denoise}. We adopt DPM-Solver++ for its efficiency and strong trajectory alignment, which better preserves colour fidelity during inversion. Nevertheless, for text-to-image diffusion models, several works\cite{hertz2022prompttoprompt,mokady2022nulltext,tumanyan2022plugandplay,wallace2022edict} report substantial reconstruction errors when applying inversion in the text-conditioned setting \(\epsilon_{\theta}(x_{t},t,t_{p})\). This issue primarily arises from the inherent instability of classifier-free guidance (CFG) mechanisms:
    \vspace{-2mm}
    {\small
    \begin{equation}
    \label{eq: cfg}
        \hat{\epsilon}_{\theta}(x_{t},t,t_{p},\varnothing )=s\cdot \epsilon_{\theta}(x_{t},t,t_{p})+(1-s)\cdot \epsilon_{\theta}(x_{t},t,\varnothing)
    \vspace{-2mm}
    \end{equation}}
    where, \(t_{p}=\mathrm{\mathcal{E}}_{t}(\textbf{T}(p))\) and \(\varnothing=\mathrm{\mathcal{E}}_{t}(`` \ ")\)
    denotes text encodings of normal and null prompt (empty string) embeddings, and \(s\) is the guidance scale. To address this, we replace null prompt with exceptional prompt embedding \(\textbf{T}_{except}\)\cite{tficon}, which sets all token numbers to a common value and removes excessive positional embeddings and special tokens like \texttt{[startoftext]}, \texttt{[endoftext]}, and \texttt{[pad]}. This leads to more accurate inversion \(\epsilon_{\theta}(x_{t},t,\mathrm{\mathcal{E}}_{t}(\textbf{T}_{except}))\), as any information included in the prompt causes backward ODE trajectories to deviate from forward trajectories.

\begin{figure}[hbt!]
    \centering
        \includegraphics[width=\linewidth]{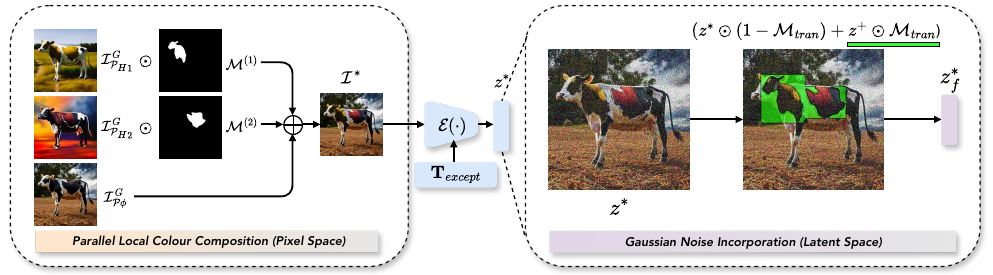}
        \vspace{-6mm}
    \caption{Initial Gaussian noise incorporation in latent space.}
    \label{fig:noise}
    \vspace{-2mm}
\end{figure}

\vspace{2mm}
    \noindent \textbf{Initial Gaussian Noise Incorporation:} After inversion process \(\mathcal{I}^{\ast}\rightarrow z^{\ast}\), we incorporate the additional Gaussian noise \(z^{+}\) to enhance transition smoothness by allowing the rich generative priors \cite{ldm} to effectively in-paint the transition area between regions inside (\ie, \(\mathcal{M}_{tot}\)) and outside (\ie, \(1-\mathcal{M}_{tot}\)) the user masks. Initially, we determine the spatial extent of each regional mask \(\mathcal{M}\) by computing its coverage rectangular area \(\mathcal{M}_{rect}\). Subsequently, we identify the transitional area \(\mathcal{M}_{tran}\) by employing the summation of XOR operations \(\oplus\) between each rectangular area and its corresponding user-defined mask as \({\mathcal{M}}_{tran}=\sum_{i=1}^{n}({{\mathcal{M}}_{rect}^{i}}\oplus{\mathcal{M}}^{i})\). Therefore, the initial noise incorporation for final diffusion ODE solving becomes: \( z^{\ast}_{f} = z^{\ast}\odot(1-\mathcal{M}_{tran})+z^{+}\odot\mathcal{M}_{tran}\) (see Fig.\ref{fig:noise}).

    \vspace{-3mm}


    \noindent \textbf{Self-attention Injection for Global Consistency:} In the final stage of latent composition, the initial noise \({z}^{\ast}_{f}\), derived from the composited image \({\mathcal{I}}^{\ast}\), serves as the starting point for solving the diffusion ODE from \(T\) to \(0\) \cite{song2021scorebased}. Throughout this stage, sketch class semantics along with additional prompt \(p\) (\eg, ``\textit{hyper-realistic} \texttt{[class]} \textit{in photography style.}'') is instrumental in producing smooth colour transitions across boundary regions. Nonetheless, relying solely on the initial noise may not sufficiently preserve the desired colour consistency (\ie, \(\textit{colour}(\mathcal{I}^{L}\odot\mathcal{M}) \not\approx\textit{colour}(\mathcal{P_{H}})\)) or maintain sketch identity (\ie, \(id(\mathcal{I}^{L}) \not\approx id(\mathcal{S})\)). To address this, we inject the self-attention map  \({\mathcal{A}}^{\ast}_{l,t}\) of \({\mathcal{I}}^{\ast}\) to guide the denoising steps. This self-attention map is derived using self-attention modules from the pretrained Stable Diffusion model \cite{ldm}, incorporating exceptional prompt embedding \(\textbf{T}_{except}\) for enhanced contextual understanding. Particularly, the self-attention module at layer \(l\) composes of three projection matrices as \(W_{Q_{l}}, W_{K_{l}}, W_{V_{l}}\) and the encoded features of composited image at timestep \(t\) and layer \(l\) is \({z}^{\ast}_{l,t}\in \mathbb{R}^{h\times w\times d}\), we compute the self-attention map \({\mathcal{A}}^{\ast}_{l,t}\)  by  \( {\mathcal{A}}^{\ast}_{l,t} = \text{softmax}\left( \frac{Q_{l,t}K^{T}_{l,t}}{\sqrt{d}} \right)\cdot V_{l,t} \) with:
    \vspace{-2mm}
    {\small
    \begin{equation}
    Q_{l,t}=W_{Q_{l}}\cdot {z}^{\ast}_{l,t},\, 
    K_{l,t}=W_{K_{l}}\cdot {z}^{\ast}_{l,t},\, 
    V_{l,t}=W_{V_{l}}\cdot {z}^{\ast}_{l,t}.
    \vspace{-2mm}
    \end{equation}}
    \vspace{-0mm}
However, achieving a balanced trade-off between textual and spatial guidance is essential. Therefore, we introduce a scaling factor \(\tau\) to regulate this trade-off. During the initial phase \(t\in[T,T\cdot(1-\tau)]\) we inject the self-attention map \({\mathcal{A}}^{\ast}_{l,t}\) to maintain \textit{global fidelity}; whereas, in the subsequent phase \(t\in[T\cdot(1-\tau),0]\) the text encoding \(t_{p}\) from the text prompt \(p\) facilitates \textit{smooth transitions} and \textit{colour-blending} throughout local regions of the final colourised result \(\mathcal{I}^{L}\).

\vspace{-1mm}
\section{Experiments}
\label{sec:experiments}
\vspace{-1mm}

    \noindent \textbf{Dataset:} We evaluate our method on four diverse datasets chosen to cover distinct domains: AFHQ-cat and AFHQ-dog \cite{choi2020stargan} for animal face images, Place365-small \cite{zhou2017places} for complex indoor and outdoor scenes, Danbooru2023\footnote{https://huggingface.co/datasets/nyanko7/danbooru2023} for anime-style content, and PascalVOC2012 \cite{pascal} for multi-object natural scenes. Specifically, we use $\sim$5k images each from AFHQ-cat/dog training sets, 900 test images from five randomly selected indoor and outdoor categories in Place365-small, the first 10k samples of Danbooru2023, and $\sim$12k images from PascalVOC2012. To obtain high-quality sketches, we combine the edge detection algorithms PiDiNet \cite{su2021pixel} and LineartDetector \cite{diff4}. Experiments are divided into \textit{global} and \textit{local} sketch colourisation. For global stage, we use K-Means clustering with $K{=}4$ to extract dominant colours (choice of $K$ detailed in Sec.~\ref{subsec:ablation}). For local stage, two random masks are applied—sufficient as mask count has no impact on results (see Sec.~\ref{subsec:local})—and dominant colours are computed within masked regions using the same $K{=}4$. We further test our method in-the-wild on 15 randomly sourced sketches from \url{www.freepik.com} using the query {\textit{``black-and-white} \texttt{[class]} \textit{sketch}''}.

\vspace{2mm}
    \noindent \textbf{Implementation Details:} We use the official Stable Diffusion v1.5 implementation \cite{ldm} with Huggingface versions of Scribble ControlNet \cite{diff4} and BLIP-2’s Q-former \cite{blip2} for the global stage, following default settings and \(p\) described in Sec.~\ref{subsec:global}. Class semantics are derived from BLIP-2’s VQA features, and colour names are retrieved via a K-D Tree \cite{kdtree} search over the RGB space restricted to the CSS3 colour database \cite{css3-colours}. The local stage employs official Stable Diffusion v1.5 \cite{ldm}, DPM-Solver++ \cite{lu2023dpmsolver}, TF-ICON \cite{tficon}, and Prompt-to-Prompt \cite{hertz2022prompttoprompt} implementations with CFG = 2.5, $\tau = 0.4$, and \(p\) described in Sec.~\ref{subsec:local}. All experiments are conducted on a single Nvidia RTX 4090 Super GPU.

\vspace{2mm}
\noindent \textbf{Evaluation Metrics:} We evaluate performance using five metrics. \textit{(i)} \textit{Fr\'echet Inception Distance (FID)} \cite{heusel2018gans} measures the distributional difference between real and generated samples using InceptionV3 activations \cite{szegedy2015inception}, where lower scores indicate better fidelity. \textit{(ii)} \textit{Learned Perceptual Image Patch Similarity (LPIPS)} \cite{zhang2018unreasonable} computes the weighted \(l2\) distance between deep features from an ImageNet-pretrained AlexNet \cite{alexnet}, with lower values implying closer perceptual alignment. \textit{(iii)} \textit{Peak Signal-to-Noise Ratio (PSNR)} compares the maximum signal power to noise distortion, where higher values reflect superior reconstruction quality. \textit{(iv)} \textit{SSIM} measures luminance, contrast, and structural similarity, with scores closer to 1 denoting higher consistency. \textit{(v)} \textit{Dynamic Closest Colour Warping (DCCW)} \cite{dccw} measures palette similarity by estimating the perceptual distance between predicted and reference colour palettes, providing accurate reflection of colour consistency.


\noindent \textbf{Competitors:} We evaluate our framework against SOTA methods for global and local sketch colourisation: (i) For \emph{global} colourisation, \textbf{DiffBlender} \cite{kim2023diffblender} enhances text-to-image diffusion with sketch and palette inputs, while \textbf{T2I-Adapter} \cite{t2iadapter} serves as a ControlNet-based baseline \cite{diff4} using identical sketches and textual prompts. \textbf{IDeepColor} \cite{zhang2017ideepcolor} is adapted by colourising grayscale T2I-Adapter outputs with 100 sampled hints from ground-truths. \textbf{Anime-Painter} \cite{anime-painter} and \textbf{CounterfeitXL} \cite{counterfeitxl}, tuned for anime styles, are tested on Danbooru2023 to assess domain generalisation. (ii) For \emph{local} colourisation, we compare with \textbf{ColorizeDiffusion} \cite{yan2024colorizediffusion}, \textbf{ColorFlow} \cite{zhuang2025colorflow}, \textbf{MangaNinja} \cite{liu2025manganinja}, and \textbf{Cobra} \cite{zhuang2025cobra}. All models receive identical reference patches; baselines use them directly, while our method extracts dominant colours and processes them with our two-stage pipeline (Sec.~\ref{sec:proposed}) to ensure fairness and consistency.

\begin{figure}[!hbt]
\centering
\vspace{-3mm}
\includegraphics[width=\linewidth]{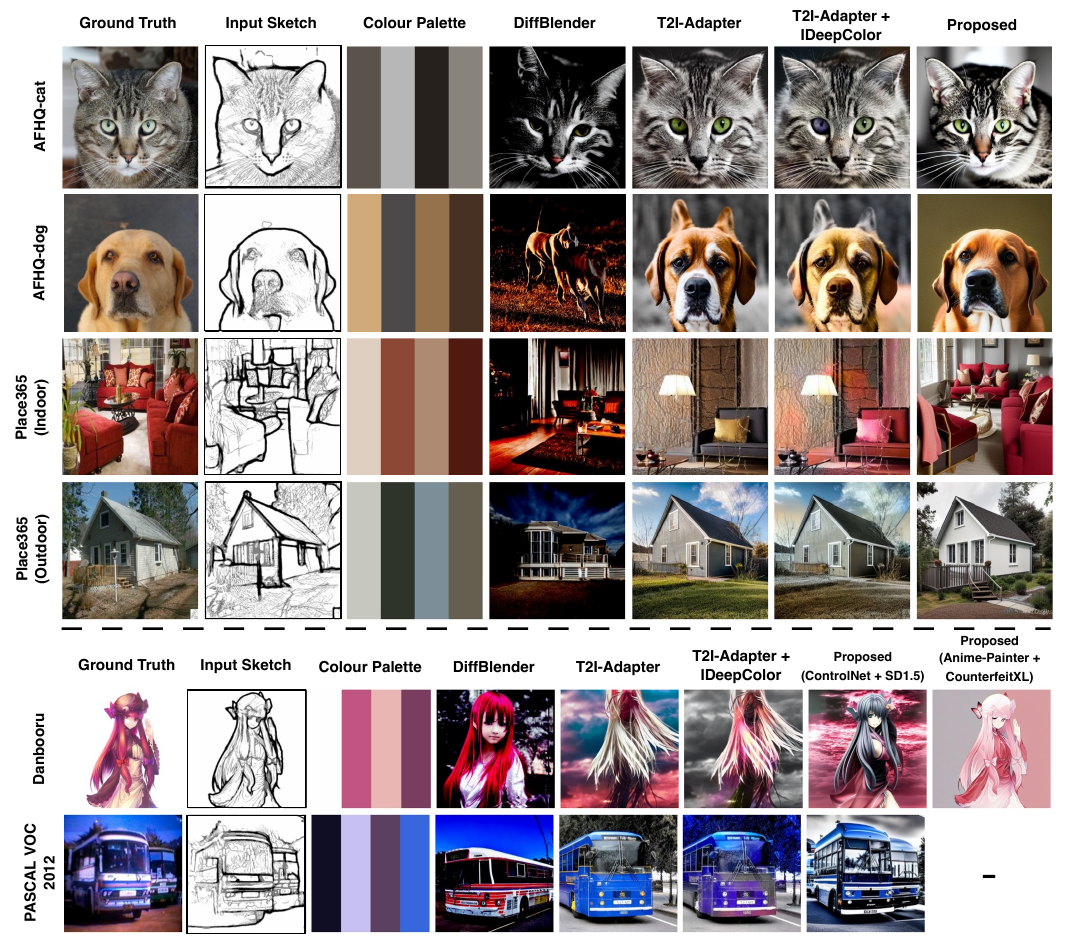}
\vspace{-6mm}
\caption{\textbf{Qualitative results on diverse datasets (\textit{global}).} Our method outperforms SOTA techniques, showing better sketch fidelity, colour vividness, realism, and overall colourisation quality.}
\label{fig:qualitative}
\vspace{-4mm}
\end{figure}

\begin{figure*}[!t]
\centering
\vspace{-8mm}
\includegraphics[width=.95\linewidth]{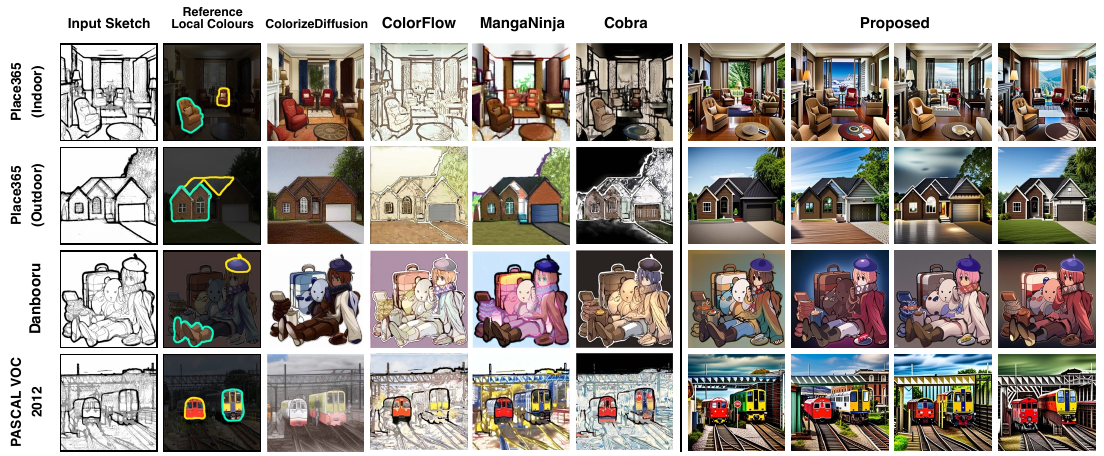}
\vspace{-2mm}
\caption{\textbf{Qualitative evaluation on diverse datasets. (\textit{local})} Our method outperforms SOTA techniques, achieving more accurate local colour propagation, and enhanced sketch fidelity. By effectively integrating local reference cues through the proposed two-stage pipeline (see Sec.~\ref{sec:proposed}), it produces coherent and realistic colourisation with improved consistency and visual quality across diverse styles and datasets.}
\label{fig:local_results}

\vspace{-3mm}

\captionof{table}{Benchmarks on diverse datasets (\textit{local colourisation}).}
\label{table:local}
\small
\renewcommand{\arraystretch}{1.3}
\setlength{\tabcolsep}{4pt}
\scalebox{0.58}{\begin{tabular}{cccccc|ccccc|ccccc|ccccc}
\toprule
\multirow{2}{*}{\textbf{Methods}} & \multicolumn{5}{c}{\textbf{Place365 (Indoor)}} & \multicolumn{5}{c}{\textbf{Place365 (Outdoor)}} & \multicolumn{5}{c}{\textbf{Danbooru2023}} & \multicolumn{5}{c}{\textbf{PascalVOC2012}} \\ 
& \textbf{FID\(\downarrow\)} & \textbf{LPIPS\(\downarrow\)} & \textbf{PSNR $\uparrow$} & \textbf{SSIM $\uparrow$} & \textbf{DCCW $\downarrow$} & \textbf{FID\(\downarrow\)} & \textbf{LPIPS\(\downarrow\)} & \textbf{PSNR $\uparrow$} & \textbf{SSIM $\uparrow$} & \textbf{DCCW $\downarrow$} & \textbf{FID\(\downarrow\)} & \textbf{LPIPS\(\downarrow\)} & \textbf{PSNR $\uparrow$} & \textbf{SSIM $\uparrow$} & \textbf{DCCW $\downarrow$} & \textbf{FID\(\downarrow\)} & \textbf{LPIPS\(\downarrow\)} & \textbf{PSNR $\uparrow$} & \textbf{SSIM $\uparrow$} & \textbf{DCCW $\downarrow$}\\ 
\addlinespace[0.5pt]
\hline
\addlinespace[1pt]

ColorizeDiffusion \cite{yan2024colorizediffusion} & 151.52 & 0.645 & 11.18 & 0.310 & 15.30 & 125.936 & 0.554 & 12.04 & 0.386 & 9.73 & 229.44 & 0.455 & 9.56 & 0.485 & 9.65 & 110.80 & 0.517 & \cellcolor{MidnightBlue!15}\textbf{11.42} & 0.257 & 24.37\\

ColorFlow \cite{zhuang2025colorflow} & 354.07 & 0.643 & 5.97 & 0.225 & 17.05 & 204.45 & 0.603 & 8.13 & 0.361 & 16.05 & 94.87 & \cellcolor{MidnightBlue!15}\textbf{0.313} & \cellcolor{MidnightBlue!15}\textbf{10.02} & \cellcolor{MidnightBlue!15}\textbf{0.541} & 10.23 & 367.69 & 0.491 & 8.14 & 0.295 & 14.98\\

MangaNinja \cite{liu2025manganinja} & 134.57 & 0.548 & 9.57 & 0.331 & 15.19 & 127.70 & 0.568 & 11.24 & 0.403 & 9.46 & 304.17 & 0.507 & 9.57 & 0.485 & 19.61 & 289.21 & 0.492 & 9.83 & 0.302 & 10.61\\

Cobra \cite{zhuang2025cobra} & 221.38 & 0.603 & 8.01 & 0.186 & 14.96 & 193.58 & 0.732 & 4.18 & 0.107 & 9.05 & 138.48 & 0.373 &6.27 & 0.402 & 13.57 & 382.70 & 0.554 & 5.72 & 0.164 & 13.96\\

 
\cellcolor{MidnightBlue!15}\textbf{Proposed} & \cellcolor{MidnightBlue!15}\textbf{123.87} & \cellcolor{MidnightBlue!15}\textbf{0.527} & \cellcolor{MidnightBlue!15}\textbf{11.65} & \cellcolor{MidnightBlue!15}\textbf{0.352} & \cellcolor{MidnightBlue!15}\textbf{11.85} & \cellcolor{MidnightBlue!15}\textbf{112.48} & \cellcolor{MidnightBlue!15}\textbf{0.489} & \cellcolor{MidnightBlue!15}\textbf{12.51} & \cellcolor{MidnightBlue!15}\textbf{0.408} & \cellcolor{MidnightBlue!15}\textbf{7.65} & \cellcolor{MidnightBlue!15}\textbf{134.54} & 0.450 & 7.31 & 0.416 & \cellcolor{MidnightBlue!15}\textbf{9.23} & \cellcolor{MidnightBlue!15}\textbf{95.64} & \cellcolor{MidnightBlue!15}\textbf{0.481} & 8.97 & \cellcolor{MidnightBlue!15}\textbf{0.298} & \cellcolor{MidnightBlue!15}\textbf{8.89}\\

\bottomrule
\end{tabular}}

\vspace{-4mm}
\end{figure*}

\vspace{-3mm}
\subsection{Qualitative Evaluation}
\vspace{-1.5mm}


\noindent \textbf{Local Colourisation:} Compared with reference-based methods, our approach yields more realistic results with vivid colours and stronger fidelity to input sketch and reference local colours (see Fig.~\ref{fig:local_results}). While MangaNinja~\cite{liu2025manganinja} transfers local colours effectively, it lacks global consistency. On Danbooru2023 dataset, Cobra~\cite{zhuang2025cobra} provides convincing texture and shading but produces less vibrant backgrounds. Thus, the results demonstrate the robustness of our \textit{training-free} framework, achieving competitive or superior local colourisation even against domain-specific methods.

\vspace{1.5mm}
\noindent \textbf{Global Colourisation:} 
Fig.~\ref{fig:qualitative} demonstrates the effectiveness of our global colourisation framework in preserving sketch structure and colour consistency across all diverse domains, from simple animal sketches (AFHQ \cite{choi2020stargan}) to complex multi-object scenes (PascalVOC2012 \cite{pascal}). Our method achieves higher sketch fidelity than competing approaches. While IDeepColor \cite{zhang2017ideepcolor} attains accurate colour placement using colour hints from ground-truth images, it still exhibits colour artefacts such as colour bleeding and desaturation, suggesting that image colourisation techniques are suboptimal for improving sketch colourisation pipeline. In contrast, our training-free design \textit{integrates seamlessly with domain-specific backbones} (\ie, Anime-Painter \cite{anime-painter} and CounterfeitXL \cite{counterfeitxl}), showing strong adaptability across domains, as evidenced on the Danbooru2023 results.

\begin{table}[h]
\begin{center}
\captionsetup{font=scriptsize}
\caption{Benchmarks on diverse datasets (\textit{global colourisation}).}
\tiny
\renewcommand{\arraystretch}{1.3}
\setlength{\tabcolsep}{1pt}
\vspace{-3mm}

\scalebox{0.9}{
\begin{tabular}{cccccc|ccccc}
\toprule
\multirow{2}{*}{\textbf{Methods}} & \multicolumn{5}{c}{\textbf{AFHQ-cat}} & \multicolumn{5}{c}{\textbf{AFHQ-dog}} \\ 
& \textbf{FID\(\downarrow\)} & \textbf{LPIPS\(\downarrow\)} & \textbf{PSNR $\uparrow$} & \textbf{SSIM $\uparrow$} & \textbf{DCCW $\downarrow$} & \textbf{FID\(\downarrow\)} & \textbf{LPIPS\(\downarrow\)} & \textbf{PSNR $\uparrow$} & \textbf{SSIM $\uparrow$} & \textbf{DCCW $\downarrow$}\\ 
\addlinespace[0.5pt]
\hline
\addlinespace[1pt]

DiffBlender \cite{kim2023diffblender} & 86.82 & 0.811 & 6.94 & 0.032 & 11.60 & 145.50 & 0.858 & 6.81 & 0.033 & 11.74\\

T2I-Adapter \cite{t2iadapter} & 68.95 & 0.706 & 7.80 & 0.134 & 10.44 & 107.12 & 0.746 & 8.12 & 0.134 & 11.66\\

T2I-Adapter + IDeepColor \cite{zhang2017ideepcolor} & 68.41 & 0.673 & 7.89 & 0.133 & \cellcolor{MidnightBlue!15}\textbf{6.93} & 116.95 & 0.699 & 8.27 & 0.135 & \cellcolor{MidnightBlue!15}\textbf{7.66}\\

 
\cellcolor{MidnightBlue!15}\textbf{Proposed} & \cellcolor{MidnightBlue!15}\textbf{50.31} & \cellcolor{MidnightBlue!15}\textbf{0.671} & \cellcolor{MidnightBlue!15}\textbf{8.65} & \cellcolor{MidnightBlue!15}\textbf{0.187} & 10.30 & \cellcolor{MidnightBlue!15}\textbf{89.70} & \cellcolor{MidnightBlue!15}\textbf{0.687} & \cellcolor{MidnightBlue!15}\textbf{8.78} & \cellcolor{MidnightBlue!15}\textbf{0.214} & 11.44\\


\end{tabular}
}

\scalebox{0.9}{
\begin{tabular}{cccccc|ccccc}
\toprule
\multirow{2}{*}{\textbf{Methods}} & \multicolumn{5}{c}{\textbf{Place365 (Indoor)}} & \multicolumn{5}{c}{\textbf{Place365 (Outdoor)}} \\ 
& \textbf{FID\(\downarrow\)} & \textbf{LPIPS\(\downarrow\)} & \textbf{PSNR $\uparrow$} & \textbf{SSIM $\uparrow$} & \textbf{DCCW $\downarrow$} & \textbf{FID\(\downarrow\)} & \textbf{LPIPS\(\downarrow\)} & \textbf{PSNR $\uparrow$} & \textbf{SSIM $\uparrow$} & \textbf{DCCW $\downarrow$}\\ 
\addlinespace[0.5pt]
\hline
\addlinespace[1pt]

DiffBlender \cite{kim2023diffblender} & 133.12 & 0.720 & 6.59 & 0.045 & 10.24 & 144.37 & 0.724 & 7.00 & 0.050 & 13.72\\

T2I-Adapter \cite{t2iadapter} & 121.48 & 0.643 & 8.80 & 0.154 & 10.81 & 176.12 & 0.740 & 9.22 & 0.100 & 13.16\\

T2I-Adapter + IDeepColor \cite{zhang2017ideepcolor} & 121.57 & 0.614 & 8.86 & 0.153 & \cellcolor{MidnightBlue!15}\textbf{7.28} & 176.40 & 0.690 & 9.42 & 0.119 & \cellcolor{MidnightBlue!15}\textbf{8.61}\\

 
\cellcolor{MidnightBlue!15}\textbf{Proposed} & \cellcolor{MidnightBlue!15}\textbf{113.99} & \cellcolor{MidnightBlue!15}\textbf{0.553} & \cellcolor{MidnightBlue!15}\textbf{9.03} & \cellcolor{MidnightBlue!15}\textbf{0.199} & 10.69 & \cellcolor{MidnightBlue!15}\textbf{129.11} & \cellcolor{MidnightBlue!15}\textbf{0.605} & \cellcolor{MidnightBlue!15}\textbf{9.46} & \cellcolor{MidnightBlue!15}\textbf{0.166} & 12.11\\

\end{tabular}
}

\scalebox{0.76}{
\begin{tabular}{cccccc|ccccc}
\toprule
\multirow{2}{*}{\textbf{Methods}} & \multicolumn{5}{c}{\textbf{Danbooru2023}} & \multicolumn{5}{c}{\textbf{PascalVOC2012}} \\ 
& \textbf{FID\(\downarrow\)} & \textbf{LPIPS\(\downarrow\)} & \textbf{PSNR $\uparrow$} & \textbf{SSIM $\uparrow$} & \textbf{DCCW $\downarrow$} & \textbf{FID\(\downarrow\)} & \textbf{LPIPS\(\downarrow\)} & \textbf{PSNR $\uparrow$} & \textbf{SSIM $\uparrow$} & \textbf{DCCW $\downarrow$}\\ 
\addlinespace[0.5pt]
\hline
\addlinespace[1pt]

DiffBlender \cite{kim2023diffblender} & 232.01 & 0.779 & 4.06 & 0.068 & 18.09 & 82.15 & 0.776 & 6.43 & 0.071 & 14.16\\

T2I-Adapter \cite{t2iadapter} & 232.70 & 0.735 & 6.57 & 0.148 & 17.11 & 154.30 & 0.697 & 8.61 & 0.155 & 11.77\\

T2I-Adapter + IDeepColor \cite{zhang2017ideepcolor} & 233.36 & 0.709 & 6.77 & 0.152 & \cellcolor{MidnightBlue!15}\textbf{13.15} & 160.62 & 0.670 & 8.74 & 0.156 & \cellcolor{MidnightBlue!15}\textbf{8.17}\\


\textbf{\cellcolor{MidnightBlue!15}Proposed (ControlNet \cite{diff4} + SD1.5 \cite{ldm})} & 149.66 & 0.657 & 6.90 & 0.237 & 15.71 & \cellcolor{MidnightBlue!15}\textbf{75.27} & \cellcolor{MidnightBlue!15}\textbf{0.558} & \cellcolor{MidnightBlue!15}\textbf{8.97} & \cellcolor{MidnightBlue!15}\textbf{0.163} & 11.12\\

\textbf{\cellcolor{MidnightBlue!15}Proposed (Anime-Painter \cite{anime-painter} + CounterfeitXL \cite{counterfeitxl})} & \cellcolor{MidnightBlue!15}\textbf{138.74} & \cellcolor{MidnightBlue!15}\textbf{0.483} & \cellcolor{MidnightBlue!15}\textbf{8.82} & \cellcolor{MidnightBlue!15}\textbf{0.438} & 15.10 & - & - & - & - & -\\

\bottomrule

\end{tabular}
}
\vspace{-8mm}
\label{table:global}
\end{center}
\end{table}

\subsection{Quantitative Evaluation}
\vspace{-1mm}

\noindent \textbf{Local Colourisation:} As shown in Table~\ref{table:local}, our method shows superior performance for indoor and outdoor scenes (see Fig.~\ref{fig:local_results}), demonstrating stronger generalisation to real-world settings. On Danbooru2023, other methods achieve better LPIPS, PSNR, and SSIM, reflecting their capabilities in the line-art domain, yet our lower FID and higher DCCW indicate greater realism and local colour fidelity. For PascalVOC2012, while methods like ColorizeDiffusion \cite{yan2024colorizediffusion} score well on LPIPS, PSNR, and SSIM, likely due to its resemblance to line-art scenes, their outputs appear less coherent with visually unappealing backgrounds (see Fig.~\ref{fig:local_results}). Overall, our approach achieves the best FID and DCCW scores across all datasets, confirming its \textit{strong global consistency} while facilitating \textit{precise local colour transfer}.

\vspace{2.5mm}

\noindent \textbf{Global Colourisation:} As shown in Table~\ref{table:global}, our method consistently outperforms competing approaches across scenarios, including animal faces and indoor/outdoor settings. Although IDeepColor \cite{zhang2017ideepcolor} attains superior DCCW performance due to its direct use of ground-truth colour hints, it degrades perceptual quality with noticeable colour artefacts (see Fig.~\ref{fig:qualitative}). On Danbooru2023, the benefits of our \textit{training-free} design are evident, offering strong domain adaptability and seamless integration of stronger pretrained backbones, such as transitioning from SD1.5 to SDXL (\ie, CounterfeitXL \cite{counterfeitxl}). Using SDXL further boosts PSNR and SSIM scores, confirming improved pixel-level fidelity for \({\mathcal{I}}^{G}_{{\mathcal{P_{H}}}}\).

\begin{figure*}[!t]
\centering
\vspace{-8mm}
\includegraphics[width=0.98\linewidth]{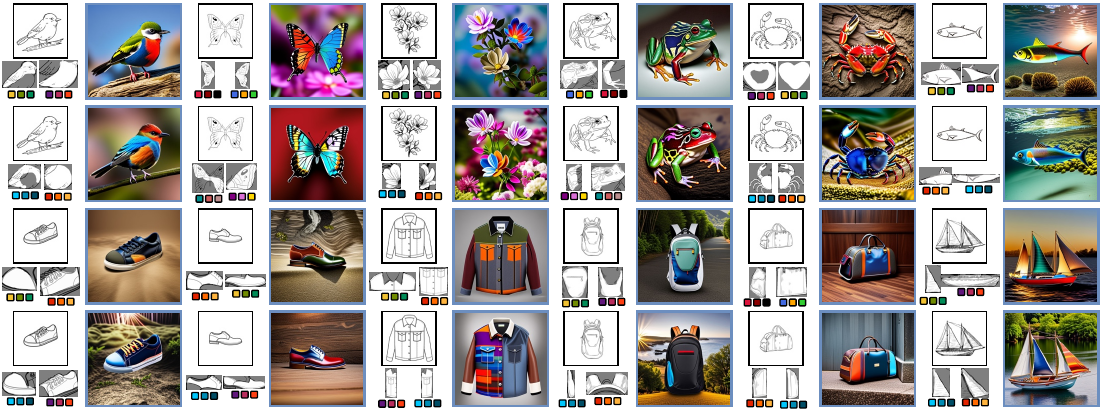}
\vspace{-2mm}
\caption{\textbf{Qualitative evaluation on \textit{in-the-wild} sketches.} We randomly selected sketches from \url{www.freepik.com} using the keywords ``\textit{black-and-white} \texttt{[class]} \textit{sketch}.'' Our sketch colourisation technique allows users to specify colour palettes and corresponding region masks (using Photoshop), with generative priors embedded in diffusion model determining the best way to apply the chosen palettes.}
\label{fig:local_results2}
\vspace{-3mm}
\end{figure*}

\subsection{Ablation Study}
\label{subsec:ablation}
\vspace{-1mm}

\noindent \textbf{Impact of colour palette size:} We evaluate the influence of palette size \(\mathcal{P}_{H}\) on the globally colourised result \({\mathcal{I}}^{G}_{\mathcal{P}_{H}}\) using LPIPS \cite{zhang2018unreasonable} for perceptual quality and DCCW \cite{dccw} for colour fidelity. As shown in Table~\ref{table:palette}, larger palettes improve colour faithfulness by capturing greater chromatic diversity, but excessive sizes introduce textual ambiguity in the T2I diffusion model, slightly reducing structural coherence, as indicated by higher LPIPS when \(n\) increases from 4 to 5.


\begin{table}[h]
\begin{center}
\captionsetup{font=scriptsize}
\caption{Ablation on different colour palette sizes.}
\tiny
\renewcommand{\arraystretch}{1.1}
\setlength{\tabcolsep}{4pt}
\vspace{-3mm}
\scalebox{.95}{
\begin{tabular}{ccc|cc|cc|cc}

\toprule
\multirow{2}{*}{\textbf{Methods}} & \multicolumn{2}{c}{\textbf{AFHQ-cat}} & \multicolumn{2}{c}{\textbf{Place365 (Indoor)}} & \multicolumn{2}{c}{\textbf{Danbooru2023}} & \multicolumn{2}{c}{\textbf{PascalVOC2012}} \\ 
& \textbf{LPIPS\(\downarrow\)} & \textbf{DCCW\(\downarrow\)} & \textbf{LPIPS\(\downarrow\)} & \textbf{DCCW\(\downarrow\)} & \textbf{LPIPS\(\downarrow\)} & \textbf{DCCW\(\downarrow\)} & \textbf{LPIPS\(\downarrow\)} & \textbf{DCCW\(\downarrow\)} \\

\addlinespace[0.5pt]
\hline
\addlinespace[1pt]

w/ 1 colour & 0.693 & 14.23 & 0.562 & 15.72 & 0.694 & 32.16 & 0.673 & 15.06 \\

w/ 2 colours & 0.674 & 12.12 & 0.559 & 12.32 & 0.676 & 19.10 & 0.676 & 12.36 \\

w/ 3 colours & 0.686 & 10.61 & 0.557 & 11.21 & 0.671 & 16.66 & 0.665 & 11.20 \\

\cellcolor{MidnightBlue!15}\textbf{w/ 4 colours} & \cellcolor{MidnightBlue!15}\textbf{0.671} & 10.30 & \cellcolor{MidnightBlue!15}\textbf{0.553} & 10.69 & \cellcolor{MidnightBlue!15}\textbf{0.657} & 15.71 & \cellcolor{MidnightBlue!15}\textbf{0.558} & 11.12 \\

\cellcolor{MidnightBlue!15}\textbf{w/ 5 colours} & 0.682 & \cellcolor{MidnightBlue!15}\textbf{9.21} & 0.560 & \cellcolor{MidnightBlue!15}\textbf{10.22} & 0.668 & \cellcolor{MidnightBlue!15}\textbf{15.55} & 0.662 & \cellcolor{MidnightBlue!15}\textbf{10.58} \\

\bottomrule
\end{tabular}
}
\label{table:palette}
\vspace{-5mm}
\end{center}
\end{table}

\begin{figure}[h]
    \centering
    \vspace{-2mm}
        \includegraphics[width=.98\linewidth]{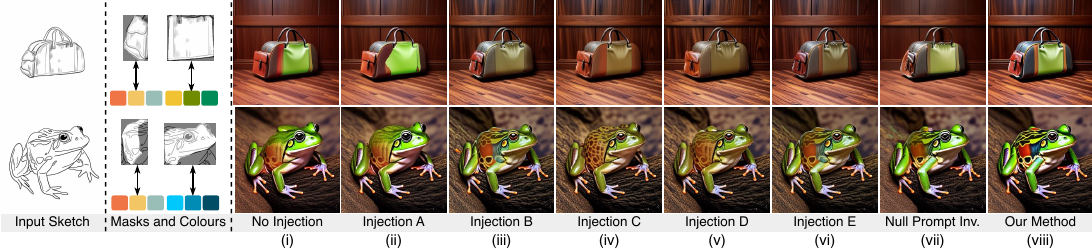}
    \vspace{-2mm}
    \caption{\textbf{Contributions of Attention Maps.\ } \textbf{\textit{(i)} No injection} (without attention map injection), \textbf{\textit{(ii-vi)} Injection A-E} (different combinations of attention map injection: 
    \textbf{A} = \textit{SA}(\(\scriptstyle {\mathcal{I}}^{\ast}\))+\textit{CA}(\(\scriptstyle{{\mathcal{I}}}^{\ast}\),
    \(\scriptstyle{\mathcal{I}}^{G}_{\mathcal{M}}\)),
    \\\textbf{B} = \textit{SA}(\(\scriptstyle{\mathcal{I}}^{\ast}\))+\textit{SA}(\(\scriptstyle{\mathcal{I}}^{G}_{\mathcal{M}}\)),
    \textbf{C} = \textit{SA}(\(\scriptstyle{\mathcal{I}}^{\ast}\))+\textit{CA}(\(\scriptstyle{\mathcal{I}}^{\ast}\), \(\scriptstyle{\mathcal{I}}^{G}_{\mathcal{M}}\))+\textit{SA}(\(\scriptstyle{\mathcal{I}}^{G}_{\mathcal{M}}\)),
    \\\textbf{D} = \textit{SA}(\(\scriptstyle{\mathcal{I}}^{G}_{\mathcal{M}}\))+\textit{CA}(\(\scriptstyle{\mathcal{I}}^{G}_{\mathcal{M}}\),
    \(\scriptstyle{\mathcal{I}}^{G}_{\mathcal{P}\phi}\)+\textit{SA}(\(\scriptstyle{\mathcal{I}}^{G}_{\mathcal{P}\phi}\)),
    \textbf{E} = \textit{SA}(\(\scriptstyle{\mathcal{I}}^{G}_{\mathcal{M}}\))+\textit{CA}(\(\scriptstyle{\mathcal{I}}^{G}_{\mathcal{M}}\),
    \(\scriptstyle{\mathcal{I}}^{G}_{\mathcal{P}\phi}\)), (\textit{SA}: self-attention and
    \textit{CA}: cross-attention)), \textbf{\textit{(vii)} Null inversion} (change from exceptional prompt to null prompt inversion \cite{mokady2022nulltext}),  \textbf{\textit{(viii)} Our method} (default settings as in \cref{subsec:local}).
    }
    \label{fig:variants}
\vspace{-2mm}
\end{figure}

\noindent \textbf{Comparison of Attention Map Injections:} Fig.~\ref{fig:variants} compares results from different attention map injection strategies. The baseline solves the diffusion ODE from $T$ to $0$ using DPM-Solver++~\cite{lu2023dpmsolver} without injections (Eq.~\ref{eq:ode}). \textit{(i)} Increasing injected maps does not improve quality, especially for local colourisation, where colour and sketch fidelity are critical. \textit{(ii)} Injecting multiple maps adds unnecessary computational cost, as each requires solving another diffusion ODE. Hence, we inject only the self-attention map of the composited image, fixing the diffusion process count at two. This design achieves better detail preservation, while global consistency is maintained through textual guidance.


\noindent \textbf{Effect of Exceptional Prompt \(\textbf{T}_{except}\):} To assess \(\textbf{T}_{except}\), we replace it with a null prompt (\ie, the unconditional term in classifier-free guidance (CFG), Eq.~\ref{eq: cfg}) during inversion. As shown in Fig.~\ref{fig:variants}, this results in reduced colour richness and content fidelity. The degradation arises from special tokens in the null prompt (\eg, \texttt{[startoftext], [endoftext]}, $\dots$) that cause embedding shifts and distort spatial coherence. \(\textbf{T}_{except}\), with these tokens removed, yields a more stable and accurate inversion, maintaining alignment between forward and backward ODE trajectories.


\begin{figure}[h]
\centering
\includegraphics[width=0.98\linewidth]{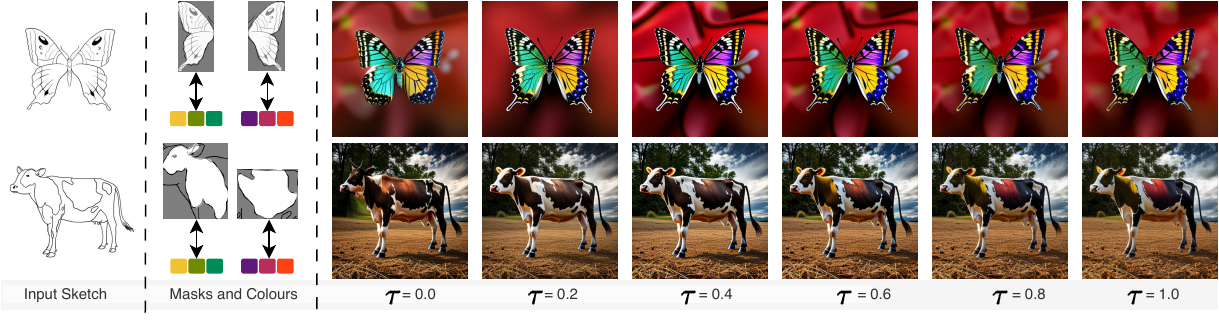}
\vspace{-3mm}
\caption{\textbf{Trade-off between harmonisation and faithfulness.} A low $\tau=0.0$ gives better harmonisation of colours but lacks faithfulness (\eg, missing regions in the wings of `butterfly') whereas a high $\tau=1.0$ gives highly faithful image generation but lacks harmonisation of colours (\eg, the red patch on ``cow'').}
\label{fig:scaling_value}
\vspace{-6mm}
\end{figure}


\vspace{3mm}
    \noindent \textbf{Benefit of Scaling Factor \(\tau\):} Fig.~\ref{fig:scaling_value} shows the effect of varying scaling factor \(\tau\). Lower \(\tau\) values yield smoother colour transitions guided mainly by the prompt (\eg, \texttt{[class]}, \textit{hyper-realistic}), but at \(\tau = 0\) the output loses alignment with the sketch and colours. Increasing \(\tau\) strengthens structural and colour fidelity via the self-attention injection (Sec.~\ref{subsec:local}), yet excessive values reduce blending and make results resemble \({\mathcal{I}}^{\ast}\). Optimal coherence occurs for \(\tau \in [0.3, 0.6]\); we set \(\tau = 0.4\) in all experiments.


\vspace{-2mm}
\section{Conclusion}
\vspace{-2mm}

\label{sec:conclusion}
In conclusion, this paper introduces a novel approach to sketch colourisation that expertly balances precision and user convenience. Our method leverages easily created Photoshop region masks and user-defined colour palettes to offer intuitive control. We utilise staged generation, latent-space composition via ODE inversion and guided sampling, and a custom self-attention mechanism with an innovative scaling factor. Our approach addresses existing limitations, such as \textit{limited user creativity} from restrictive textual prompts, inaccurate local colour assignment, and unpleasant results. Our design, which requires no training or fine-tuning, enhances zero-shot capabilities and integrates smoothly with more advanced foundational methods. Finally, our model is fast (takes less than $20$ steps), training-free, and works on consumer-grade Nvidia RTX 4090 GPU.

\section*{Acknowledgements}
We acknowledge the support of UK Research and Innovation (UKRI) through the UKRI AI: Centre for Doctoral Training in AI for Digital Media Inclusion under Grant No.: EP/Y030915/1. For the purpose of open access, the author has applied a Creative Commons Attribution (CC BY) license to any Author Accepted Manuscript version arising.


{
    \small
    \bibliographystyle{ieeenat_fullname}
    \bibliography{main}
}

\newpage
\twocolumn[
\begin{center}
    {\Large Supplementary Material for\\
    \textbf{SketchDeco: Training-Free Latent Composition for Precise Sketch Colourisation}}
\end{center}
]
\input{supp}



\end{document}

%% file: supp.tex
\def\paperID{} 
\def\confName{CVPR}
\def\confYear{2026}


\renewcommand\thesection{\Alph{section}}


\section{Supplementary Material}
\label{sec:supplement}
\vspace{-1mm}

\noindent This supplementary material complements the main paper, "SketchDeco: Training-Free Latent Composition for Precise Sketch Colourisation," by providing additional details: ablation study of textual prompts (Sec.~\ref{sec:ablative}), the unusual effectiveness of K-D Tree (Sec.~\ref{sec:benefit}), ablation on local sketch colourisation (Sec.~\ref{sec:local_abla}), additional discussion on design choices (Sec.~\ref{sec:design}), clarification on contributions (Sec.~\ref{sec:contribution}), limitation and future study (Sec.~\ref{sec:limitation}), and full quantitative results on in-the-wild sketch dataset (Sec.~\ref{sec:qualitative}).

\setcounter{figure}{0}
\renewcommand{\figurename}{Fig.}
\renewcommand{\thefigure}{S\arabic{figure}}

\setcounter{table}{0}
\renewcommand{\tablename}{Table}
\renewcommand{\thetable}{S\arabic{table}}

\vspace{-2mm}
\section{Ablation Study of Textual Prompts}
\label{sec:ablative}
\vspace{-2mm}

\begin{figure}[h]
\centering
\vspace{-3mm}
\includegraphics[width=\linewidth, trim=0.8cm 0cm 0.8cm 0cm, clip]{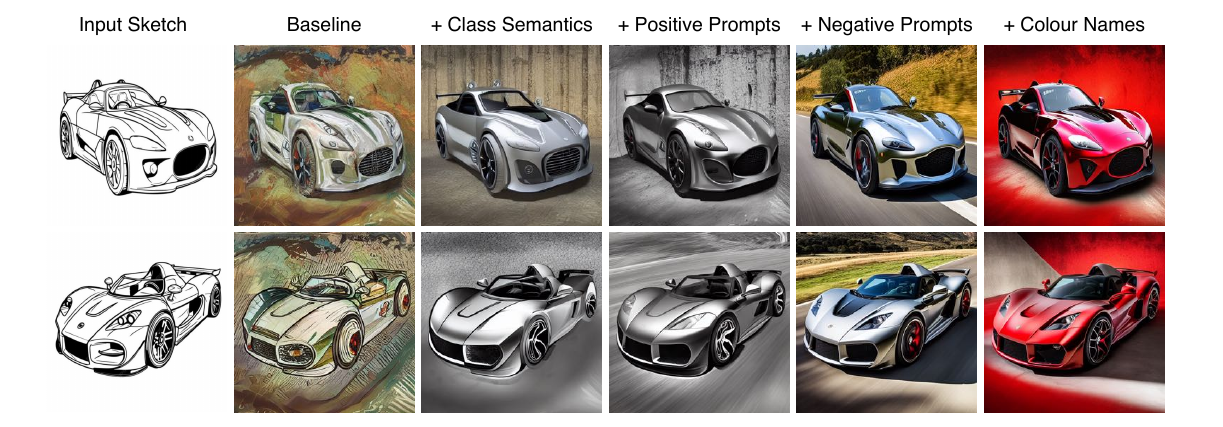}
\vspace{-7mm}
\caption{\textbf{Ablation study of different textual prompt settings.} In our global sketch colourisation pipeline, the textual prompts $t_{p}$ are composed of four main textual information, including class semantics, colour names, positive prompts (i.e., ``\texttt{[class]}\textit{, hyper-realistic, quality, photography style, using only colours in colour palette of} \texttt{[$c_s^1$]}, - \texttt{[$c_s^2$]} - $\dots$ \texttt{[$c_s^n$]}'') and negative prompts (i.e., ``\textit{drawing look, sketch look, line art style, cartoon look, unnatural colour, unnatural texture, unrealistic look, low-quality}'')}
\label{fig:ablative}
\vspace{-2mm}
\end{figure}

We conduct an additional ablation study to investigate the contribution of each textual information on our global sketch colourisation process in the following cases: \textit{(i)} Baseline, where the colourised results are generated without any textual guidance (i.e., only input sketches are utilised as conditions); \textit{(ii)} only class semantics (i.e., ``\texttt{[class]}''), derived from BLIP-2\cite{blip2} prediction, are applied for textual guidance; \textit{(iii)} class semantics and positive prompts (i.e., ``\texttt{[class]}\textit{, hyper-realistic, quality, photography style}'') are used; \textit{(iv)} negative prompts are combined; \textit{(v)} finally, four main textual information, including class semantics, colour names, positive prompts (i.e., ``\texttt{[class]}\textit{, hyper-realistic, quality, photography style, using only colours in colour palette of} \texttt{[$c_s^1$]}, - \texttt{[$c_s^2$]} - $\dots$ \texttt{[$c_s^n$]}'') and negative prompts (i.e., ``\textit{drawing look, sketch look, line art style, cartoon look, unnatural colour, unnatural texture, unrealistic look, low-quality}'') are utilised. The experimental outcomes demonstrate that each component of textual guidance plays a pivotal role in generating faithful result. Notably, the presence of colour names and class semantics significantly enhances the visual outputs, as depicted in Fig. \ref{fig:ablative}.

\section{The Unusual Effectiveness of K-D Tree}
\label{sec:benefit}
\vspace{-2mm}
\noindent \textbf{[i] Compatability with professional designer workflow:} In our envisioned user interface (UI), we prioritise convenience by allowing designers to specify region masks and their desired colour palettes using Photoshop. While these palettes are typically provided as colour codes (\eg, \texttt{\#FFD700} or \texttt{rgb(255,215,0)}), our diffusion models require text-based input to specify colours (\ie, ``\texttt{gold}''). To bridge this gap between user-friendly inputs and the requirements of Stable Diffusion \cite{diff4}, we propose a straightforward yet highly effective solution -- a binary search tree (specifically, a K-D Tree with $K=3$) that maps RGB values $\mathcal{P}_{rgb}=\left\{(r, g, b)\;|\;r, g, b \in [0, 255]\right\}$ to a W3C standard database of $147$ CSS3 colour names. 

\begin{figure}[h]
\centering
\vspace{-4mm}
\includegraphics[width=.95\linewidth]{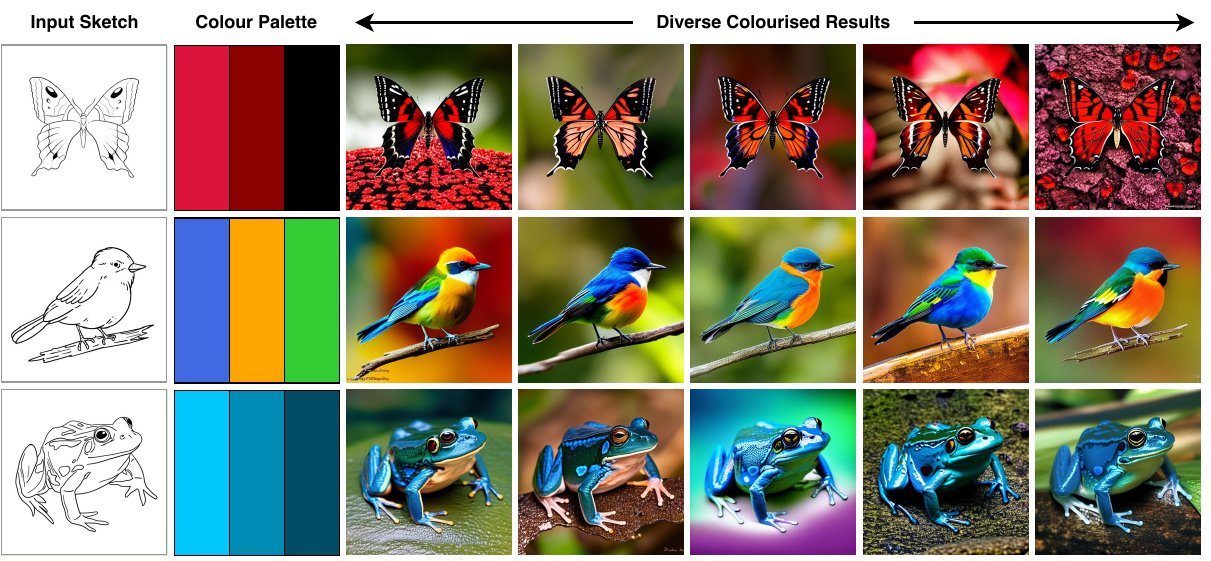}
\vspace{-2mm}
\caption{\textbf{Diverse colourised outputs.} Our training-free global sketch colourisation technique leverages the rich generative priors from a large-scale text-to-image (T2I) diffusion model to produce diverse and realistic outputs. Colour names are derived from the K-D Tree algorithm with specific colours \texttt{[$c_s^1$]}, \texttt{[$c_s^2$]}, $\dots$, \texttt{[$c_s^n$]}}
\label{fig:kdtree}
\vspace{-4mm}
\end{figure}

\noindent \textbf{[ii] Choice of standard CSS3 colour database:} Remarkably, we found that using these standard CSS3 colour names as input to the pre-trained text encoder CLIP \cite{clip} effectively conditions Stable Diffusion to accurately reproduce the specified colour in the generated images, which are depicted in Fig. \ref{fig:kdtree}. This unexpected effectiveness may stem from the fact that both CLIP and Stable Diffusion were trained on large-scale internet data, where Cascading Style Sheets (CSS) are widely used as a language for describing the rendering of HTML and XML documents. By leveraging the inherent compatibility of these models with web standards, we eliminate the need for training a custom Hexcode encoder, keeping SketchDeco entirely training-free. 

\noindent \textbf{[iii] Benefits over LLMs:} When compared to LLMs, which are computationally intensive for tasks such as colour naming, a KD-tree provides a more efficient solution. LLMs are prone to generating non-existent colours, which can lead to inaccuracies in diffusion models. In contrast, a KD-tree enables quick and efficient lookups with minimal overhead. By aligning a KD-tree with standard CSS3 colours, we ensure accuracy, scalability, and easy future expansions.

\section{Ablation on Local Sketch Colourisation}
\label{sec:local_abla}

\noindent \textbf{[i] Benefits of staged generation:}
Our method employs a divide-and-conquer strategy to facilitate a complex task (see Fig.\textcolor{cvprblue}{2}). The global stage generates initial outputs adhering to predefined sketch and colour palettes. These outputs are subsequently combined in the local stage, guided by region masks, which carefully refine the results to ensure seamless colour transitions and maintain the integrity of the sketch and colours. Moreover, this design also offers user flexibility, allowing for verification and potential regeneration of global colourised results, as depicted in Fig.\ref{fig:kdtree}.

\begin{figure}[h]
\centering
\vspace{-3mm}
\includegraphics[width=.95\linewidth]{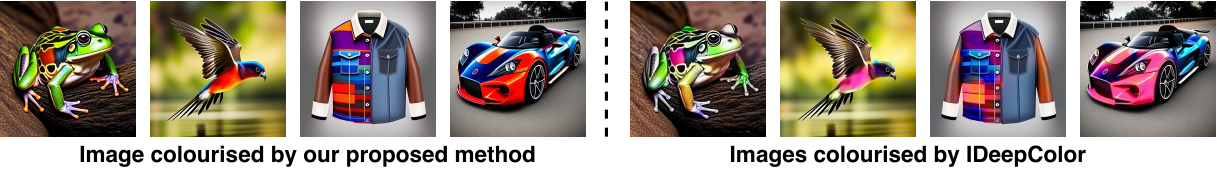}
\vspace{-3mm}
\caption{\textbf{Impact of local colour composition.} Our method can maintain colour accuracy in the local stage due to the implementation of the local colour composition step. In contrast, IDeepColour \cite{zhang2017ideepcolor} falls short and also introduces colour artefacts, e.g., grayish.}
\label{fig:local1}
\vspace{-4mm}
\end{figure}

\noindent \textbf{[ii] Choice of local colour composition}
The composition of global colourised results serves as an essential bridge between the global and local stages. This process ensures the preservation of sketch and colour integrity for $\mathcal{I}^{\ast}$. As the global outcomes are consistently derived from the same sketch, their integration is seamless. We further examine the efficacy of our approach by contrasting it with the user-guided image colourisation technique, IDeepColor \cite{zhang2017ideepcolor}, which utilises convolutional neural network (CNN) to propagate colour hints. Our experimental setup involved providing a grayscale image along with 100 random hints to IDeepColor for recolourisation. Fig. \ref{fig:local1} illustrates that IDeepColor not only fails short to retain the original colours but also introduces visual artefacts and colour bleeding.

\begin{figure}[h]
\centering
\vspace{-4mm}
\includegraphics[width=.99\linewidth]{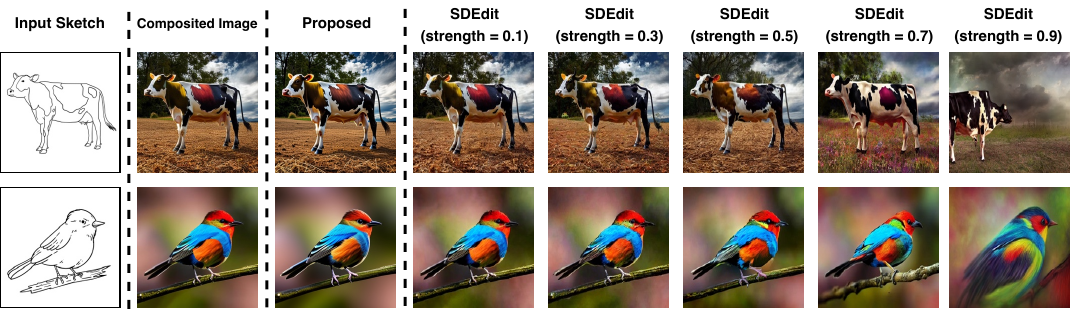}
\vspace{-2mm}
\caption{\textbf{Ablation study on local refinement steps.} Evaluation of our inversion-based local refinement technique compared to the inversion-free SDEdit\cite{meng2022sdedit} method. SDEdit tends to oversmooth images, compromising creative colour blending and texture fidelity, resulting in distorted textures (e.g., cow at \(s = 0.5\)) and unrealistic features (e.g., bird at \(s = 0.5\)). In contrast, our method retains the original colours and sketches from the composite image while facilitating harmonious colour transitions across different regions.}
\label{fig:local2}
\vspace{-4mm}
\end{figure}

\noindent \textbf{[iii] Does inversion of composited image $\mathcal{I}^{\ast}$ and local refinement steps help?:}
In this ablation study, we examine the effectiveness of our local refinement process initiated by the ODE inversion of $\mathcal{I}^{\ast}$. We compare this to SDEdit \cite{meng2022sdedit}, an inversion-free image editing method, using only Gaussian noise addition to facilitate editing process. We fixed the guidance scale at 6.5, varying only the intensity of the noise to assess the impact on the local nuances. As shown in Fig. \ref{fig:local2}, SDEdit generally smooths images but struggles with creative colour blending and detail preservation, leading to distorted textures (e.g., in a cow at \(s=0.5\)) and unrealistic features (e.g., in a bird at \(s=0.5\)). In contrast, our method surpasses SDEdit by integrating ODE inversion of $\mathcal{I}^{\ast}$ together with \(\textbf{T}_{except}\) and injection of self-attention to preserve the integrity of sketch and colour. Then, we enhance smooth colour transitions through strategic noise incorporation and textual guidance. Additionally, we also introduce the parameter \(\tau\) to effectively balance these elements during the final sampling process, facilitating an optimal trade-off between fidelity and aesthetic refinement.

\vspace{2mm}

\begin{figure}[h]
\centering
\vspace{-1mm}
\includegraphics[width=\linewidth]{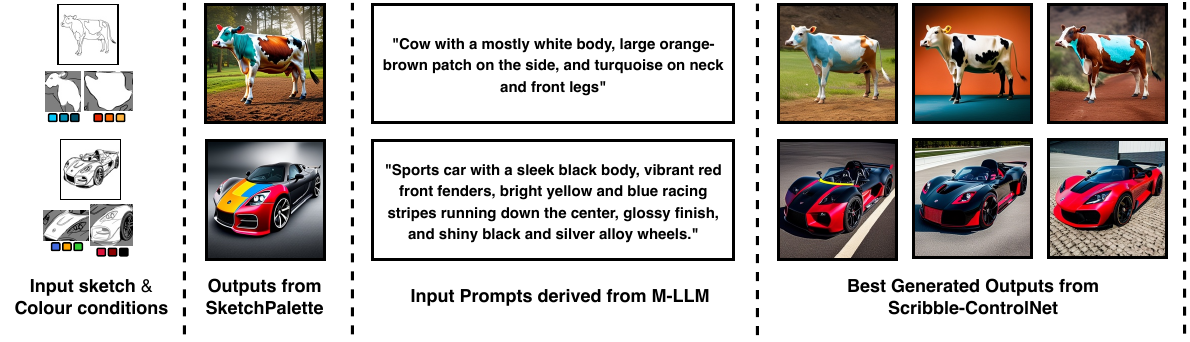}
\vspace{-7mm}
\caption{\textbf{Comparison with a naive ControlNet baseline.} We compare our method, which leverages ControlNet in global stage, against a traditional text-guided ControlNet model. The naïve ControlNet fails to reproduce our results, demonstrating its inability to enforce fine-grained, spatially localised colour constraints.}
\label{fig:sub_control}
\vspace{-4mm}
\end{figure}

\noindent \textbf{[iv] Can naive ControlNet match our fine-grained colour control?:} We additionally investigate whether a standard pretrained Scribble ControlNet~\cite{diff4}, originally developed for text-guided sketch-to-image generation, can approximate the performance of our training-free, region-controlled colourisation framework. To construct a fair comparison, captions were generated from our method's outputs \(\mathcal{I}^{\mathcal{L}}\) using multimodal LLMs, and these captions were subsequently used as text prompts for ControlNet. Even after multiple attempts and manual selection of the top three result candidates, the reproduced images remained unable to match the spatial fidelity and chromatic precision achieved by our method. This outcome indicates that text-guided conditioning is inherently limited for the task of precise local sketch colourisation, thereby reinforcing the strength of our approach in delivering fine-grained, spatially aligned colour control without any additional training (see Fig.~\ref{fig:sub_control}).

\section{Additional Discussion on Design Choices}
\label{sec:design}
In our study, we assess the impact of several design choices within the contexts of AFHQ-cat\cite{choi2020stargan} and AFHQ-dog\cite{choi2020stargan}. Firstly, within Visual Question Answering (VQA) tasks, substituting BLIP2 with GIT \cite{wang2022git} results in FID/LPIPS scores of $(65.71/0.693)$ and $(129.19/0.742)$, respectively. Comparatively, the adoption of llama-3-vision-alpha \cite{liu2023visual} yields $(57.23/0.673)$ and $(61.22/0.683)$, while llava-1.5-7b \cite{liu2023visual} delivers $(55.97/0.669)$ and $(62.96/0.686)$. Secondly, in the domain of local sketch colourisation through inversion schemes, replacing DPM-Solver++ with DDIM results in FID/LPIPS scores of $(35.09/0.716)$, with DPM-Solver achieving $(33.74/0.714)$. These findings highlight the nuanced influence of specific algorithmic changes on model performance across different tasks.

\section{Clarification on Contributions}
\label{sec:contribution}
Image/sketch colourisation is a long-standing \cite{deep5, kim2023diffblender, deep1} open-problem in Computer Vision. While prior works used GANs \cite{deep5}, the latest works \cite{kim2023diffblender} use fine-tuned diffusion models. We take the next step by introducing \textit{(i)} training-free paradigm for sketch colourisation that \textit{outperforms fine-tuned SOTA} (see Fig. \ref{fig:sota}), \textit{(ii)} existing diffusion-based colourisation only allow global colour consistency, whereas SketchDeco enable precise user-directed local colourisation, \textit{(iii)} intuitively blend several modules for a fast pipeline that is compatible with consumer grade GPUs. While the use of modules like BLIP-2, ControlNet, SD-1.5 have become ubiquitous in computer vision (due to their reliable performance) our method intuitively integrates them to introduce several ``{firsts}'' that advances image/sketch colourisation literature. Particularly, \textit{(i)} We use coarse-to-fine multi-staged generation with diffusion models to combine global and local colourisation, \textit{(ii)} Prior colourisation methods using SD-1.5 lacks local colour consistency. We adapt SD-1.5 to ensure both global and local colour consistency defined by region masks and colour palette. \textit{(iii)} A latent-space composition technique combining diffusion inversion and guided sampling to ensure both local colour fidelity and global harmony. \textit{(iv)} A custom attention mechanism that adaptively balances sketch structure preservation with controlled colour diffusion.

\section{Limitation and Future Study}
\label{sec:limitation}

In our work, we acknowledge several limitations and suggest avenues for future research. Firstly, when processing small selected regions for local colourisation, the initial blending ability derived from global colourisation pipeline may lead to some losses in colour fidelity, necessitating exploration of novel techniques to address this challenge. This could involve the development of adaptive blending strategies or localised colour correction mechanisms to better preserve colour accuracy in intricate or fine-textured areas. Additionally, while our approach relies on matching colour hexcodes with name entries in a standard CSS3 colour database\cite{css3-colours}, there is potential to enhance the database to accommodate a wider variety of colours. However, caution must be exercised to ensure compatibility with the pretrained Stable Diffusion\cite{ldm} model, as ambiguous or non-standard colour names may introduce uncertainty and inconsistency in the colourisation process. 

\section{Qualitative Results on Random Sketches}
\label{sec:qualitative}

\begin{figure}[H]
\centering
\includegraphics[width=\linewidth]{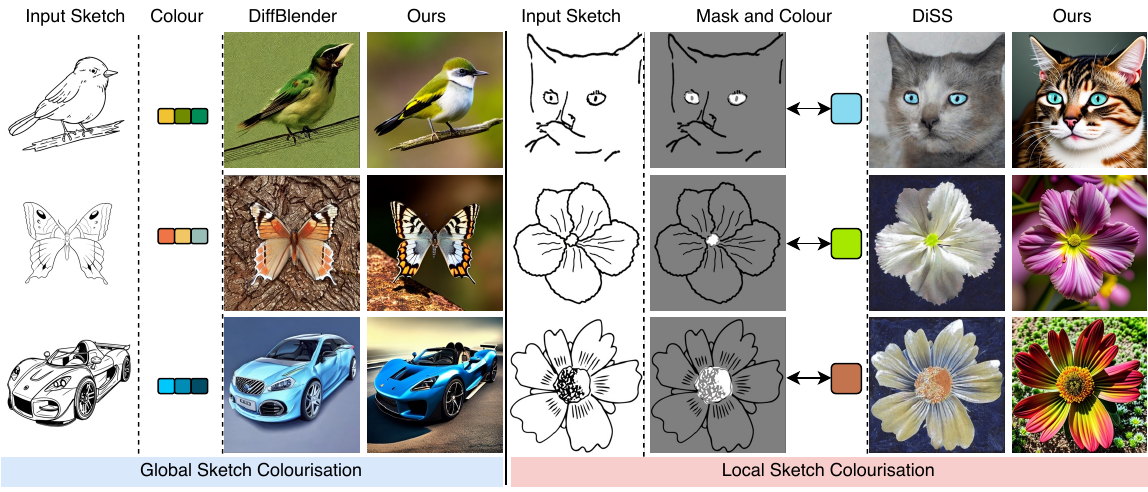}
\vspace{-3mm}
\caption{\textbf{Qualitative assessment against fine-tuned SOTA methods using randomly sourced internet sketches.} We demonstrate that our training-free approach surpasses contemporary state-of-the-art methods (i.e., DiffBlender \cite{kim2023diffblender} and DiSS \cite{diff1}) that rely on extensive fine-tuning, particularly in areas such as sketch fidelity, the application of creative colour schemes, and enhanced realism.}
\label{fig:sota}
\vspace{-4mm}
\end{figure}

\begin{figure*}[!h]
    \centering
    \vspace{-4mm}
    \resizebox{.58\linewidth}{!}{%
    \begin{tabular}{c|c|c|c}
         Input Sketch & Mask with colour & Composited Image & Final Result \\
         \hline
         \includegraphics[width=0.2\linewidth]{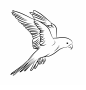} &
         \includegraphics[width=0.35\linewidth]{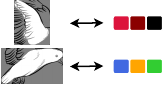} &
         \includegraphics[width=0.2\linewidth]{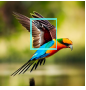} &
         \includegraphics[width=0.2\linewidth]{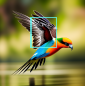} \\
         \hline
         \includegraphics[width=0.2\linewidth]{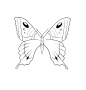} &
         \includegraphics[width=0.35\linewidth]{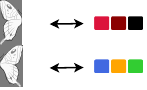} &
         \includegraphics[width=0.2\linewidth]{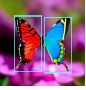} &
         \includegraphics[width=0.2\linewidth]{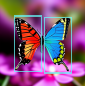} \\
         \hline
         \includegraphics[width=0.2\linewidth]{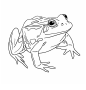} &
         \includegraphics[width=0.35\linewidth]{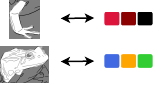} &
         \includegraphics[width=0.2\linewidth]{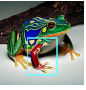} &
         \includegraphics[width=0.2\linewidth]{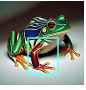} \\
         \hline
         \includegraphics[width=0.2\linewidth]{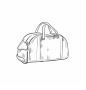} &
         \includegraphics[width=0.35\linewidth]{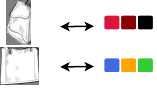} &
         \includegraphics[width=0.2\linewidth]{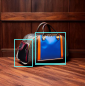} &
         \includegraphics[width=0.2\linewidth]{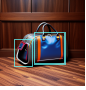} \\
         \hline
         \includegraphics[width=0.2\linewidth]{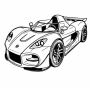} &
         \includegraphics[width=0.35\linewidth]{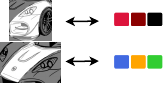} &
         \includegraphics[width=0.2\linewidth]{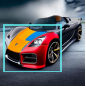} &
         \includegraphics[width=0.2\linewidth]{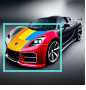} \\
         \hline
             \includegraphics[width=0.2\linewidth]{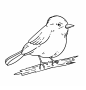} &
             \includegraphics[width=0.35\linewidth]{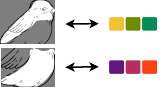} &
             \includegraphics[width=0.2\linewidth]{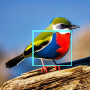} &
             \includegraphics[width=0.2\linewidth]{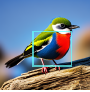} \\
             \hline
             \includegraphics[width=0.2\linewidth]{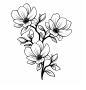} &
             \includegraphics[width=0.35\linewidth]{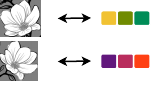} &
             \includegraphics[width=0.2\linewidth]{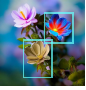} &
             \includegraphics[width=0.2\linewidth]{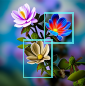} \\
             \hline
             \includegraphics[width=0.2\linewidth]{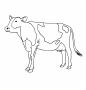} &
             \includegraphics[width=0.35\linewidth]{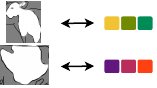} &
             \includegraphics[width=0.2\linewidth]{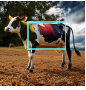} &
             \includegraphics[width=0.2\linewidth]{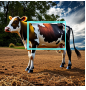} \\
             \hline
             \includegraphics[width=0.2\linewidth]{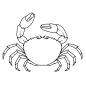} &
             \includegraphics[width=0.35\linewidth]{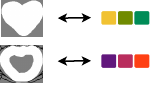} &
             \includegraphics[width=0.2\linewidth]{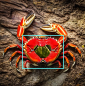} &
             \includegraphics[width=0.2\linewidth]{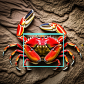} \\
             \hline
             \includegraphics[width=0.2\linewidth]{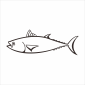} &
             \includegraphics[width=0.35\linewidth]{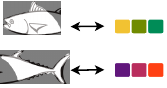} &
             \includegraphics[width=0.2\linewidth]{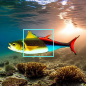} &
             \includegraphics[width=0.2\linewidth]{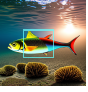} \\
    \end{tabular}
    }
    \caption{\textbf{Qualitative evaluation with \textit{in-the-wild} sketches.} The local sketch colourisation pipeline showcased notable adaptability, allowing users to incorporate three conditional inputs: a variety of sketches, hand-drawn masks, and preferred colour palettes. Consequently, our approach yields precise outcomes tailored to user-defined specifications. It is non-trivial to mention that these sketch images are randomly sourced from \url{www.freepik.com} with search keywords: ``\texttt{black-and-white [class] sketch}''.}
    \label{fig:results1}
\end{figure*}

\begin{figure*}[t]
    \centering
    \vspace{-4mm}
    \resizebox{.58\linewidth}{!}{%
    \begin{tabular}{c|c|c|c}
         Input Sketch & Mask with colour & Composited Image & Final Result \\
         \hline
         \includegraphics[width=0.2\linewidth]{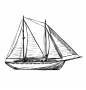} &
         \includegraphics[width=0.35\linewidth]{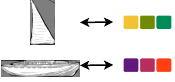} &
         \includegraphics[width=0.2\linewidth]{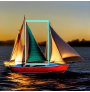} &
         \includegraphics[width=0.2\linewidth]{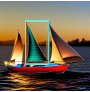} \\
         \hline
         \includegraphics[width=0.2\linewidth]{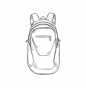} &
         \includegraphics[width=0.35\linewidth]{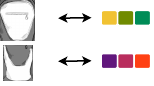} &
         \includegraphics[width=0.2\linewidth]{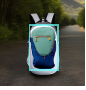} &
         \includegraphics[width=0.2\linewidth]{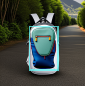} \\
         \hline
         \includegraphics[width=0.2\linewidth]{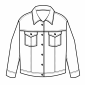} &
         \includegraphics[width=0.35\linewidth]{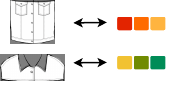} &
         \includegraphics[width=0.2\linewidth]{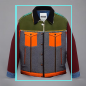} &
         \includegraphics[width=0.2\linewidth]{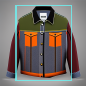} \\
         \hline
         \includegraphics[width=0.2\linewidth]{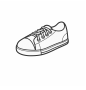} &
         \includegraphics[width=0.35\linewidth]{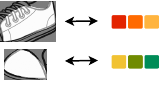} &
         \includegraphics[width=0.2\linewidth]{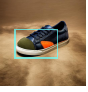} &
         \includegraphics[width=0.2\linewidth]{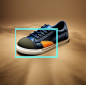} \\
         \hline
         \includegraphics[width=0.2\linewidth]{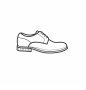} &
         \includegraphics[width=0.35\linewidth]{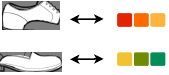} &
         \includegraphics[width=0.2\linewidth]{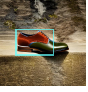} &
         \includegraphics[width=0.2\linewidth]{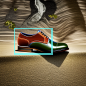} \\
         \hline
         \includegraphics[width=0.2\linewidth]{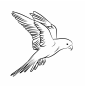} &
         \includegraphics[width=0.35\linewidth]{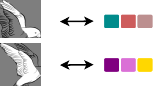} &
         \includegraphics[width=0.2\linewidth]{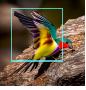} &
         \includegraphics[width=0.2\linewidth]{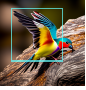} \\
         \hline
         \includegraphics[width=0.2\linewidth]{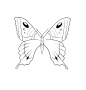} &
         \includegraphics[width=0.35\linewidth]{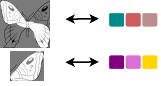} &
         \includegraphics[width=0.2\linewidth]{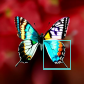} &
         \includegraphics[width=0.2\linewidth]{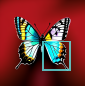} \\
         \hline
         \includegraphics[width=0.2\linewidth]{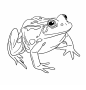} &
         \includegraphics[width=0.35\linewidth]{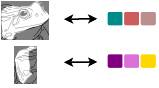} &
         \includegraphics[width=0.2\linewidth]{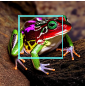} &
         \includegraphics[width=0.2\linewidth]{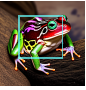} \\
         \hline
         \includegraphics[width=0.2\linewidth]{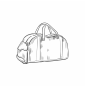} &
         \includegraphics[width=0.35\linewidth]{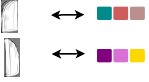} &
         \includegraphics[width=0.2\linewidth]{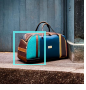} &
         \includegraphics[width=0.2\linewidth]{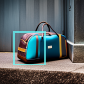} \\
         \hline
         \includegraphics[width=0.2\linewidth]{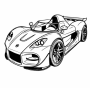} &
         \includegraphics[width=0.35\linewidth]{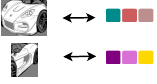} &
         \includegraphics[width=0.2\linewidth]{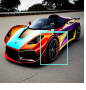} &
         \includegraphics[width=0.2\linewidth]{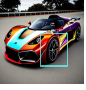} \\
    \end{tabular}
    }
    \caption{\textbf{Qualitative evaluation with \textit{in-the-wild} sketches.} The local sketch colourisation pipeline showcased notable adaptability, allowing users to incorporate three conditional inputs: a variety of sketches, hand-drawn masks, and preferred colour palettes. Consequently, our approach yields precise outcomes tailored to user-defined specifications. It is non-trivial to mention that these sketch images are randomly sourced from \url{www.freepik.com} with search keywords: ``\texttt{black-and-white [class] sketch}''.}
    \label{fig:results3}
\end{figure*}

\begin{figure*}[t]
    \centering
    \vspace{-4mm}
    \resizebox{.58\linewidth}{!}{%
    \begin{tabular}{c|c|c|c}
         Input Sketch & Mask with colour & Composited Image & Final Result \\
         \hline
         \includegraphics[width=0.2\linewidth]{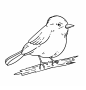} &
         \includegraphics[width=0.35\linewidth]{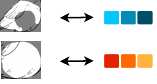} &
         \includegraphics[width=0.2\linewidth]{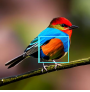} &
         \includegraphics[width=0.2\linewidth]{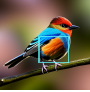} \\
         \hline
         \includegraphics[width=0.2\linewidth]{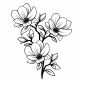} &
         \includegraphics[width=0.35\linewidth]{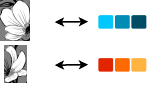} &
         \includegraphics[width=0.2\linewidth]{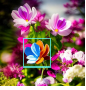} &
         \includegraphics[width=0.2\linewidth]{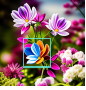} \\
         \hline
         \includegraphics[width=0.2\linewidth]{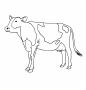} &
         \includegraphics[width=0.35\linewidth]{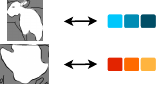} &
         \includegraphics[width=0.2\linewidth]{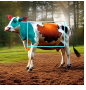} &
         \includegraphics[width=0.2\linewidth]{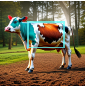} \\
         \hline
         \includegraphics[width=0.2\linewidth]{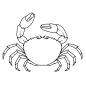} &
         \includegraphics[width=0.35\linewidth]{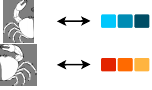} &
         \includegraphics[width=0.2\linewidth]{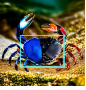} &
         \includegraphics[width=0.2\linewidth]{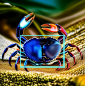} \\
         \hline
         \includegraphics[width=0.2\linewidth]{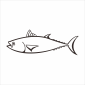} &
         \includegraphics[width=0.35\linewidth]{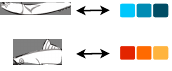} &
         \includegraphics[width=0.2\linewidth]{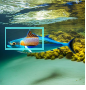} &
         \includegraphics[width=0.2\linewidth]{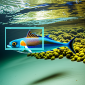} \\
         \hline
         \includegraphics[width=0.2\linewidth]{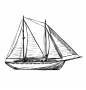} &
         \includegraphics[width=0.35\linewidth]{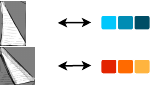} &
         \includegraphics[width=0.2\linewidth]{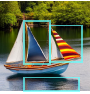} &
         \includegraphics[width=0.2\linewidth]{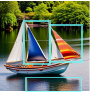} \\
         \hline
         \includegraphics[width=0.2\linewidth]{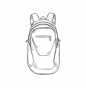} &
         \includegraphics[width=0.35\linewidth]{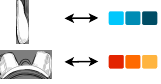} &
         \includegraphics[width=0.2\linewidth]{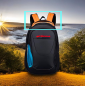} &
         \includegraphics[width=0.2\linewidth]{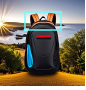} \\
         \hline
         \includegraphics[width=0.2\linewidth]{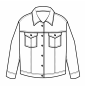} &
         \includegraphics[width=0.35\linewidth]{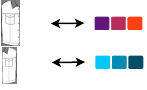} &
         \includegraphics[width=0.2\linewidth]{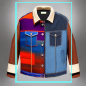} &
         \includegraphics[width=0.2\linewidth]{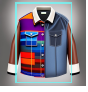} \\
         \hline
         \includegraphics[width=0.2\linewidth]{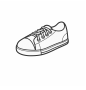} &
         \includegraphics[width=0.35\linewidth]{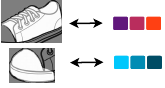} &
         \includegraphics[width=0.2\linewidth]{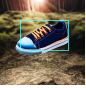} &
         \includegraphics[width=0.2\linewidth]{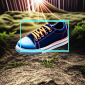} \\
         \hline
         \includegraphics[width=0.2\linewidth]{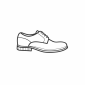} &
         \includegraphics[width=0.35\linewidth]{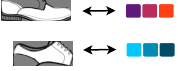} &
         \includegraphics[width=0.2\linewidth]{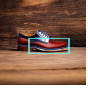} &
         \includegraphics[width=0.2\linewidth]{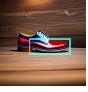} \\
    \end{tabular}
    }
    \caption{\textbf{Qualitative evaluation with \textit{in-the-wild} sketches.} The local sketch colourisation pipeline showcased notable adaptability, allowing users to incorporate three conditional inputs: a variety of sketches, hand-drawn masks, and preferred colour palettes. Consequently, our approach yields precise outcomes tailored to user-defined specifications. It is non-trivial to mention that these sketch images are randomly sourced from \url{www.freepik.com} with search keywords: ``\texttt{black-and-white [class] sketch}''.}
    \label{fig:results5}
\end{figure*}

We perform qualitative comparison against two state-of-the-art (SOTA) diffusion-based approaches: DiffBlender \cite{kim2023diffblender} and DiSS\cite{diff1}. DiffBlender aims to enhance the functionality of Text-to-Image diffusion models by integrating inputs from multiple modalities. These inputs are processed based on their conditional types, which include image/sketches, spatial tokens (\ie, bounding boxes), and non-spatial tokens (\ie, colour palettes). For fair comparison, we use a combination of sketch and colour palette modalities, equivalent to our global sketch colourisation. As for DiSS, the primary objective is to generate realistic images based on sketch and stroke conditions. DiSS also proposes a region-sensitive stroke-to-image method using partially coloured strokes as input, to synthesise diverse content in the non-coloured regions. For a fair comparison with our approach, we use a single colour instead of a colour palette. To the best of our knowledge, there is no prior work addressing the use of masks and colour palettes for \textit{region-aware colourisation}. Importantly, unlike our method, existing SOTA methods \cite{kim2023diffblender, diff1} are \textit{not training-free} and require expensive fine-tuning. As a result, from Fig. \ref{fig:sota}, our training-free colourisation approach outperforms fine-tuned state-of-the-art (SOTA) models in various aspects including colour vividness, colour harmonisation, fidelity to the original sketch, and overall realistic look. Additionally, \cref{fig:results1,fig:results3,fig:results5} represents qualitative results on 15 random sketches obtained from \url{www.freepik.com} using search keywords: {``\textit{black-and-white} \texttt{[class]} \textit{sketch}''}. Notably, the seamless colour transition effects before and after undergoing our local refinement processes in the local stage are highlighted using \textbf{\texttt{\textcolor{cvprblue}{blue bounding boxes}}}.


